\documentclass[lettersize,journal]{IEEEtran}
\usepackage{amsmath,amsfonts}
\usepackage{algorithmic}
\usepackage{array}
\usepackage[caption=false,font=normalsize,labelfont=sf,textfont=sf]{subfig}
\usepackage{textcomp}
\usepackage{stfloats}
\usepackage{url}
\usepackage{verbatim}
\usepackage{graphicx}
\def\BibTeX{{\rm B\kern-.05em{\sc i\kern-.025em b}\kern-.08em
    T\kern-.1667em\lower.7ex\hbox{E}\kern-.125emX}}
\usepackage{balance}
\usepackage{graphicx}
\usepackage{lineno,hyperref}
\usepackage{amssymb, amsmath}
\usepackage{bm}
\usepackage{color}
\usepackage{threeparttable}
\usepackage{cite}
\usepackage{color}
\usepackage{booktabs}
\usepackage{pifont}
\usepackage{array} 
\usepackage{multirow}
\usepackage{tabularx}
\usepackage{algorithm}
\usepackage{algorithmic}
\usepackage{tabularray}
\usepackage{makecell}
\usepackage{xcolor} 
\usepackage{booktabs} 

\makeatletter
\newcommand{\removelatexerror}{\let\@latex@error\@gobble}
\makeatother

\begin{document}

\title{EarthMapper: Visual Autoregressive Models for Controllable Bidirectional Satellite–Map Translation}

\author{
	Zhe~Dong,
	Yuzhe~Sun,
	Tianzhu~Liu,~\IEEEmembership{Member,~IEEE},
	Wangmeng~Zuo,~\IEEEmembership{Senior Member,~IEEE},
	and~Yanfeng~Gu,~\IEEEmembership{Senior Member,~IEEE},
	
\thanks{Z. Dong, Y. Sun, T. Liu and Y. Gu are with the School of Electronics and Information Engineering, Harbin Institute of Technology, Harbin 150001, China. (email: guyf@hit.edu.cn).}
\thanks{W. Zuo is with the Faculty of Computing, Harbin Institute of Technology, Harbin, China, and also with the Peng Cheng Lab, Shenzhen, China.}
}

\maketitle

\begin{abstract}

Satellite imagery and maps, as two fundamental data modalities in remote sensing, offer direct observations of the Earth's surface and human-interpretable geographic abstractions, respectively. The task of bidirectional translation between satellite images and maps (BSMT) holds significant potential for applications in urban planning and disaster response. However, this task presents two major challenges: first, the absence of precise pixel-wise alignment between the two modalities substantially complicates the translation process; second, it requires achieving both high-level abstraction of geographic features and high-quality visual synthesis, which further elevates the technical complexity. To address these limitations, we introduce EarthMapper, a novel autoregressive framework for controllable bidirectional satellite-map translation. EarthMapper employs geographic coordinate embeddings to anchor generation, ensuring region-specific adaptability, and leverages multi-scale feature alignment within a geo-conditioned joint scale autoregression (GJSA) process to unify bidirectional translation in a single training cycle. A semantic infusion (SI) mechanism is introduced to enhance feature-level consistency, while a key point adaptive guidance (KPAG) mechanism is proposed to dynamically balance diversity and precision during inference. We further contribute CNSatMap, a large-scale dataset comprising 302,132 precisely aligned satellite-map pairs across 38 Chinese cities, enabling robust benchmarking. Extensive experiments on CNSatMap and the New York dataset demonstrate EarthMapper’s superior performance, achieving significant improvements in visual realism, semantic consistency, and structural fidelity over state-of-the-art methods. Additionally, EarthMapper excels in zero-shot tasks like in-painting, out-painting and coordinate-conditional generation, underscoring its versatility. The source code for EarthMapper and the CNSatMap dataset will be publicly available at \href{https://github.com/HIT-SIRS/EarthMapper}{https://github.com/HIT-SIRS/EarthMapper}.

\end{abstract}


\section{Introduction}

\begin{figure}[tbp]
	\begin{center}
		\centerline{\includegraphics[width=1\linewidth]{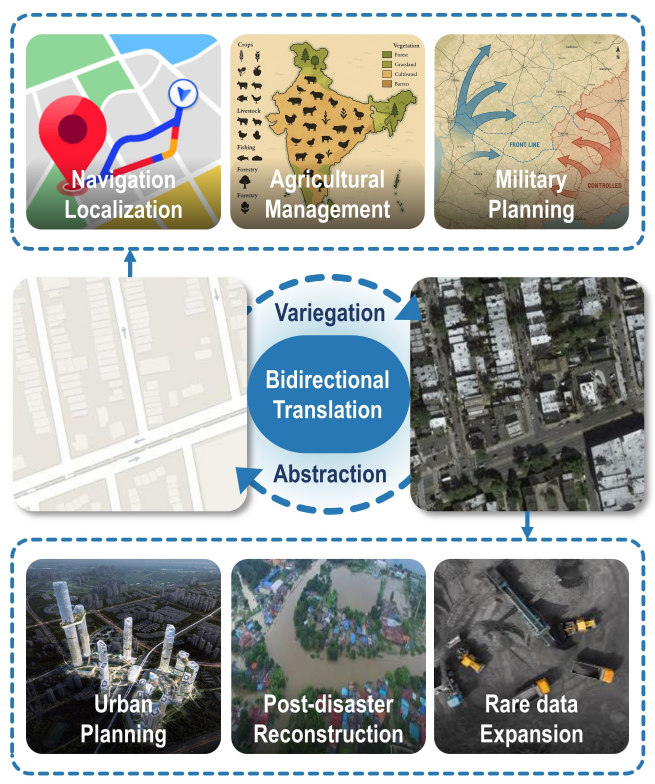}}
		\caption{Conceptual illustration of bidirectional satellite-map translation and their respective applications.}\label{introduction}
		\vspace{-10pt} 
	\end{center}
\end{figure}

\IEEEPARstart{R}emote sensing technology has emerged as a pivotal tool for acquiring geospatial information, with satellite imagery and cartographic maps serving as two primary modalities. Satellite images capture raw, unprocessed representations of the Earth's surface, while maps provide abstract, human-interpretable depictions of geographic features. In practical applications, as illustrated in Fig.~\ref{introduction}, bidirectional translation between these modalities is often indispensable: transforming satellite imagery into maps facilitates rapid comprehension of key geographic information\cite{ingale2021image}, whereas synthesizing realistic satellite images from edited maps enables rapid scenario simulations for post-disaster reconstruction, urban planning, and augmentation of satellite datasets with rare scenes\cite{espinosa2023generate}. Consequently, bidirectional satellite-map translation (BSMT) holds significant potential for advancing both remote sensing applications and automated cartography.

Recent advancements in deep learning have spurred the development of powerful generative models, achieving remarkable success in image-to-image translation tasks. Generative adversarial networks (GANs)\cite{goodfellow2020generative, mirza2014conditional} have been a cornerstone of this progress, leveraging adversarial training to refine synthetic outputs through a competitive interplay between generator and discriminator networks. Variants such as Pix2Pix\cite{isola2017image} and CycleGAN\cite{zhu2017unpaired} have demonstrated exceptional performance in style transfer and image enhancement. Meanwhile, diffusion models\cite{ho2020denoising, dhariwal2021diffusion} have introduced a probabilistic framework for high-fidelity image synthesis, progressively refining details through iterative denoising. More recently, autoregressive (AR) models\cite{chen2020generative, van2016conditional} have gained prominence by explicitly modeling pixel-level dependencies, offering superior generation quality and diversity. Notably, VAR\cite{tian2024visual} pioneered multi-scale autoregressive prediction to mitigate context fragmentation and blurriness in traditional AR approaches. Subsequent works like ControlVar\cite{li2024controlvar} and CAR\cite{yao2024car} further incorporated conditional control mechanisms, proving highly effective in cross-domain mapping tasks. These advances inspire our exploration of AR-based frameworks for high-quality BSMT. Nevertheless, dedicated research on BSMT remains scarce, necessitating further methodological innovation.

Despite the prowess of existing generative models, BSMT presents unique challenges beyond conventional image translation or style transfer. Satellite images and maps lack strict pixel-wise correspondence, as cartographic abstraction involves selective emphasis, simplification, and symbolization to enhance human interpretability. Merely performing pixel-level classification on satellite data fails to produce high-quality maps. Conversely, reconstructing realistic satellite imagery from maps demands not only textural synthesis but also semantically coherent and geographically plausible scene generation. Thus, BSMT requires models capable of both high-level abstraction and photorealistic reconstruction—a dual objective that existing methods struggle to fulfill.

To address these challenges, we propose EarthMapper, a novel autoregressive framework for BSMT with geographic coordinate embedding. First, EarthMapper initiates generation from geographic coordinates, enabling region-specific adaptation to diverse terrains and urban layouts. Second, it enforces multi-scale feature alignment between conditional inputs and generated outputs within the designed geo-conditioned joint scale autoregression (GJSA) process, permitting bidirectional translation via a single training cycle. Third, instead of direct pixel manipulation, EarthMapper modulates the autoregressive probability distribution at each step, balancing fidelity and diversity. Additionally, we introduce a semantic infusion (SI) mechanism during training, which constrains generation by minimizing feature-space discrepancies between synthetic and real images. For inference, we devise key point adaptive guidance (KPAG), comprising: (1) key point force (KPF), which anchors generation to salient geographic features in the conditional input to prevent excessive deviation; and (2) complexity guidance (CG), which dynamically adjusts the conditioning strength based on per-stage image complexity.


To propel BSMT research, we curate CNSatMap, a large-scale dataset of 302,132 precisely aligned satellite-map pairs spanning 38 major Chinese cities. Satellite images, sourced from Google Earth at 0.6-meter resolution, preserve intricate ground details, while vector maps from Tianditu service offer cartographically rigorous representations. Covering diverse climates, population densities, and urbanization levels, CNSatMap provides a robust benchmark for evaluating cross-modal translation algorithms. This dataset bridges a critical gap in BSMT research and promises broad utility in urban scene parsing, automated mapping, and geospatial feature extraction.

In summary, the contributions of this work can be summarized in the following three aspects:

\begin{itemize}

\item[(1)] We construct CNSatMap, the first large-scale, high-precision dataset for BSMT, enabling rigorous exploration of cross-modal geographic translation and fostering advancements in urban analytics and automated cartography.

\item[(2)] We propose EarthMapper, a  AR-based generative framework that unifies GJSA process via multi-scale alignment, achieving superior generative capability and versatility.

\item[(3)] The SI mechanism is designed to enforce feature-level consistency between generated and real images, enhancing semantic fidelity in weakly aligned satellite-map pairs.

\item[(4)] We introduce KPAG, which intelligently balances diversity and accuracy through key-point anchoring and dynamic complexity modulation, ensuring cartographic precision while preserving realism.

\end{itemize}

The remainder of this paper is structured as follows: Section~\ref{section:Related Work} surveys related work in controllable image generation, vision autoregressive models, and remote sensing image generation. Section~\ref{section:Dataset Construction} introduces CNSatMap dataset, detailing its construction, coverage, and quality controls. In Section~\ref{section:methods}, we describe the proposed EarthMapper in detail. Section~\ref{section:EXPERIMENTS} evaluates EarthMapper through comprehensive experiments, including comparisons with state-of-the-art methods, ablation studies, and zero-shot generalization across diverse tasks. Finally, Section~\ref{section:Conclusion} summarizes findings and future directions.

\section{Related Work}
\label{section:Related Work}

\subsection{Controllable Image Generation}

Controllable image generation (CIG) has emerged as a cornerstone in the fields of computer vision and generative modeling, offering transformative potential across diverse applications, including artistic creation, image editing, and automated content generation. Contemporary approaches to CIG can be systematically classified into three distinct paradigms based on the nature of the input conditions: label-based control, visual control, and text-based control.

Label-based control leverages structured annotations—such as semantic segmentation masks, class labels, or layout descriptors—to guide image synthesis with high precision. This paradigm excels in applications requiring strict spatial or categorical alignment. Early work, exemplified by conditional generative adversarial networks (CGANs)\cite{mirza2014conditional}, demonstrated that conditioning generators on discrete class labels enhances generation quality. Recent innovations like ControlNet\cite{zhang2023adding} advance this framework by integrating spatial cues (e.g., edge maps or human poses) into diffusion models\cite{rombach2022high} via trainable adapters, achieving fine-grained control over object placement and structure. In layout-to-image synthesis, GLIGEN\cite{li2023gligen} introduces gated self-attention layers to process bounding box information, enabling seamless compositional generation of multiple objects. These techniques often depend on paired training data—a persistent bottleneck—yet advances in multi-task learning and dataset augmentation are alleviating this constraint.

Visual control employs image-based inputs—such as sketches, edge maps, or reference images—to direct generative processes, effectively bridging abstract textual prompts and fine-grained pixel-level outcomes. Techniques like GAN inversion \cite{karras2019style} map images into latent spaces for semantic editing, with advancements like InterFaceGAN \cite{shen2020interfacegan} enabling real-time attribute manipulation. Cutting-edge brain-guided generation\cite{takagi2023high} decodes functional magnetic resonance imaging (fMRI) or electroencephalography (EEG) signals to reconstruct visual concepts, highlighting cross-modal innovation. For artistic domains, methods such as Textual Inversion \cite{gal2022image} and DreamBooth\cite{ruiz2023dreambooth} fine-tune models with minimal reference images, preserving distinct visual identities. Challenges remain in interpreting ambiguous sketches or incomplete inputs, yet hybrid strategies integrating diffusion models with attention mechanisms are enhancing robustness.

Text-based control has emerged as a cornerstone of CIG research, prized for its adeptness at encoding high-level semantics. Trailblazing models such as GLIDE\cite{nichol2021glide} and DALL·E 2\cite{ramesh2022hierarchical} solidified text-conditioned diffusion as the state-of-the-art, harnessing large-scale language models like CLIP\cite{radford2021learning} to achieve robust cross-modal alignment. Stable diffusion\cite{rombach2022high} marked a leap in efficiency by operating within a compressed latent space, facilitating high-resolution synthesis on consumer-grade hardware. Innovations like classifier-free guidance\cite{ho2022classifier} strike an elegant balance between diversity and fidelity through the joint training of conditional and unconditional diffusion models. Recent advances tackle multilingual generation and compositional reasoning, as exemplified by composable diffusion\cite{tang2023any}, though persistent challenges, such as text-image misalignment, remain.

\subsection{Vision Autoregressive Models}

Autoregressive models, renowned for their success in natural language processing (NLP) due to their ability to model long-range dependencies and generate coherent sequences\cite{radford2019language, brown2020language}, have recently emerged as powerful tools in computer vision. These models have demonstrated significant potential in tasks such as image generation\cite{parmar2018image, chen2020generative}, super-resolution\cite{li2016video, guo2022lar}, image editing\cite{yao2022outpainting, crowson2022vqgan}, and image-to-image translation\cite{li2024controlar}, driven by their sequential prediction framework. Depending on the representation strategy, autoregressive vision models can be broadly categorized into pixel-based, token-based, and scale-based approaches.

Pixel-based autoregressive models, such as PixelRNN\cite{van2016pixel} and PixelCNN\cite{van2016conditional}, treat images as 1D sequences of pixels, predicting each pixel conditioned on previous ones using architectures like LSTMs or CNNs. While effective for low-resolution image generation, these models struggle with high-resolution tasks due to quadratic computational complexity and inherent redundancy in pixel-level predictions. Although parallelization techniques\cite{reed2017parallel} have been explored, the generated images often suffer from blurriness and suboptimal quality, highlighting the limitations of pixel-based approaches for scaling to higher resolutions.

Token-based autoregressive models address these limitations by leveraging discrete token representations. Frameworks like VQ-VAE\cite{van2017neural}, VQ-VAE-2\cite{razavi2019generating}, and VQGAN\cite{esser2021taming} compress images into sequences of discrete tokens using vector quantization (VQ), enabling efficient processing of high-resolution content. These models typically employ a two-stage process: an encoder-decoder architecture learns discrete latent representations, followed by an autoregressive model that predicts token sequences for generation. By integrating Transformer-based decoders\cite{vaswani2017attention} and perceptual losses, token-based models achieve state-of-the-art performance in high-resolution image generation while maintaining computational efficiency.

Scale-based autoregressive models, such as VAR\cite{tian2024visual}, introduce a hierarchical approach to image generation by processing visual content across multiple scales, from coarse to fine. Unlike token-based models that predict tokens sequentially, VAR generates entire token maps at each scale using residual quantization (RQ), as introduced in RQ-VAE\cite{lee2022autoregressive}. This recursive quantization of feature residuals allows for compact representation of high-resolution images while preserving fine details. The multi-scale framework enables parallel token generation within each map, improving spatial locality and computational efficiency. 

\subsection{Remote Sensing Image Generation}

Remote sensing image generation has become a critical task in Earth observation, addressing the growing need for synthetic images that accurately mimic those captured by satellites or unmanned aerial vehicles (UAVs). The scarcity and high cost of acquiring high-resolution annotated data, coupled with the variability introduced by factors such as acquisition time, weather conditions, and sensor types, make this task both challenging and essential. Recent advancements in generative models have shown significant promise in overcoming these challenges, enabling the synthesis of high-quality remote sensing images for applications ranging from environmental monitoring and urban planning to agricultural assessment.

Among these advancements, HSIGene\cite{pang2024hsigene} stands out as a foundation model for hyperspectral image generation. By leveraging latent diffusion models (LDMs), HSIGene can produce precise hyperspectral images with detailed spectral information, which is critical for applications such as crop health monitoring and environmental analysis. Besides, RSDiff\cite{sebaq2024rsdiff} and CRS-Diff\cite{tang2024crs} have introduced multi-stage diffusion processes to generate high-resolution satellite imagery from text prompts. RSDiff employs a two-stage framework, combining a low-resolution diffusion model (LRDM) with a super-resolution diffusion model (SRDM) to enhance spatial details, while CRS-Diff incorporates multi-scale feature fusion to improve control over conditional inputs such as text, metadata, and reference images. These models excel in handling multispectral and time-series data, offering superior geographic detail and resolution compared to traditional methods. Further expanding the scope of generative capabilities, MetaEarth\cite{yu2024metaearth} introduces a resolution-guided self-cascading framework, enabling the generation of unbounded, large-scale remote sensing images tailored to specific geographic and resolution requirements. This approach is particularly valuable for global-scale applications such as climate modeling and urban development planning. The field has also been significantly advanced by DiffusionSat \cite{khanna2023diffusionsat}, which supports diverse conditional generation tasks such as environmental monitoring and crop yield prediction, demonstrating the versatility and scalability of diffusion models in Earth observation. Despite these impressive advancements, challenges remain in achieving precise conditional control and ensuring the stability of generated images in complex scenarios.

\begin{figure*}[tbp]
	\begin{center}
		\centerline{\includegraphics[width=1\linewidth]{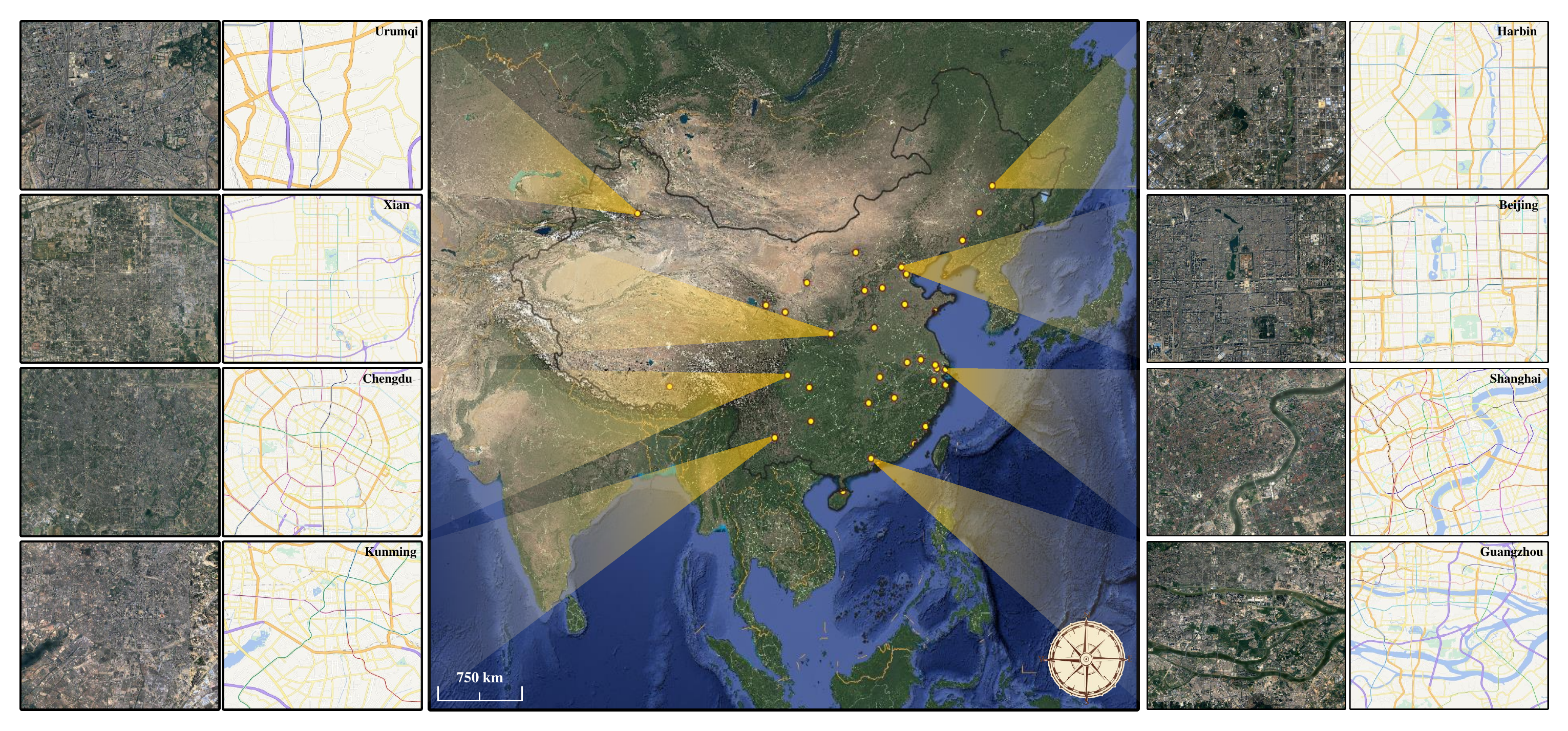}}
		\caption{Illustration of the geographical distribution of satellite-map pairs sampled from the proposed CNSatMap dataset.}\label{dataset}
		\vspace{-10pt} 
	\end{center}
\end{figure*}

\section{The CNSatMap Dataset}
\label{section:Dataset Construction}

In this study, we introduce the CNSatMap dataset, a comprehensive resource designed for cross-modal translation tasks between remote sensing satellite imagery and cartographic maps. CNSatMap comprises 302,132 meticulously aligned satellite-map image pairs, making it the largest publicly available dataset of its kind. It is tailored to support geospatial analysis, cross-modal image translation, and urban scene understanding across diverse geographical contexts. The dataset was constructed through a rigorous pipeline to ensure high quality, precise alignment, and semantic richness, as detailed below.

The satellite imagery in CNSatMap is sourced from Google Earth, captured at zoom level 19 with a spatial resolution of approximately 0.6 meters per pixel. This resolution preserves fine-grained spatial details essential for applications such as urban planning, infrastructure inspection, and land-use classification. For cartographic data, we exclusively utilize the Tianditu map service, a vector-based mapping platform maintained by China’s National Geomatics Center. This choice was motivated by the limitations of alternative services, such as Google Maps, which exhibit infrequent updates in Chinese regions—often lagging by months or years—and insufficient coverage of localized geographical features. In contrast, Tianditu provides up-to-date, high-precision cartographic data with exceptional fidelity in administrative boundaries, terrain representation, and urban infrastructure details.

To ensure broad geographical representativeness, CNSatMap encompasses 38 major Chinese cities, including national capitals, provincial capitals, municipalities, and first-tier cities: Beijing, Changchun, Changsha, Chengdu, Chongqing, Dalian, Fuzhou, Guangzhou, Guiyang, Haikou, Hangzhou, Harbin, Hefei, Hohhot, Jinan, Kunming, Lanzhou, Lhasa, Nanchang, Nanjing, Ningbo, Qingdao, Shanghai, Shenyang, Shenzhen, Shijiazhuang, Suzhou, Taiyuan, Tianjin, Wuhan, Urumqi, Wuxi, Xiamen, Xi’an, Hong Kong, Xining, Yinchuan, and Zhengzhou. These cities span diverse climatic zones, population densities, and urbanization levels. Coastal metropolises like Shanghai and Shenzhen exemplify highly developed urban agglomerations, while inland cities such as Lhasa and Urumqi reflect unique topographical and sparsely populated characteristics. This distribution, illustrated in Fig.~\ref{dataset}, underscores the dataset’s socioeconomic and geospatial diversity.

\begin{figure*}[tbp]
	\begin{center}
		\centerline{\includegraphics[width=1\linewidth]{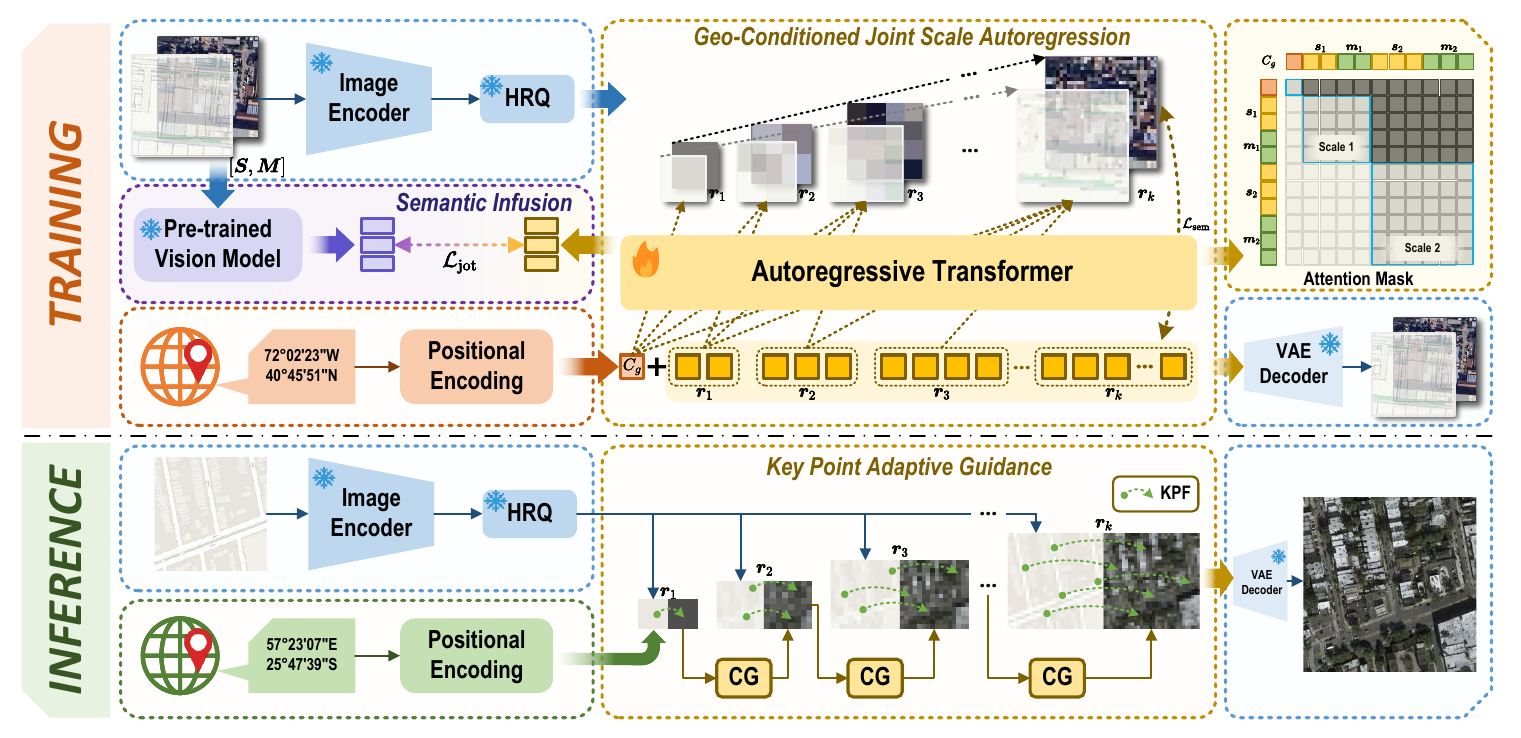}}
		\caption{Overview of our proposed EarthMapper framework, with the upper section dedicated to training and the lower section to inference. During training, paired satellite and map data are initially processed by a frozen image encoder and quantized into vector representations using the hierarchical residual quantization (HRQ) module. Concurrently, geographic coordinates are encoded as initial vectors and fed into the geo-conditioned joint scale autoregression (GJSA) model, which generates image content by progressively increasing the resolution. The resulting output is then passed to the semantic infusion (SI) module, where it is aligned with ground truth features to improve the realism of the generated image, before being decoded into an image using the VAE decoder. In the inference phase, for instance, when generating satellite images from maps, the map is quantized into vectors, combined with geographic coordinate vectors, and input into the model. Key points are computed using the key point force (KPF) method and incorporated into the generation process, while complexity guidance (CG) dynamically modulates the control intensity. Finally, the output is decoded to produce the satellite image.}\label{flowchart}
	\end{center}
\end{figure*}

The construction of CNSatMap involved a multi-stage workflow to ensure alignment accuracy and semantic coherence. Raw satellite and map imagery for the 38 cities were collected and georeferenced using the EPSG:3857 (Web Mercator) projection, the standard for web mapping services. Each image was partitioned into non-overlapping 256 $\times$ 256-pixel tiles, balancing computational efficiency with contextual preservation—a common input size for vision generation models in computer vision. Tile boundary coordinates were then transformed from EPSG:3857 to EPSG:4326 (WGS84 latitude-longitude), ensuring compatibility with geospatial information systems (GIS) through embedded metadata. A geometric alignment process was applied to correct potential misregistrations caused by discrepancies in data sources or georeferencing errors, ensuring pixel-level correspondence between satellite and map modalities.

From an initial pool of over 600,000 tiles, three filtering criteria were enforced to refine the dataset. First, satellite tiles obscured by cloud cover or atmospheric artifacts were removed. Second, tiles with significant building tilt due to non-orthogonal imaging angles were excluded. Finally, map tiles with low structural diversity (standard deviation of color distribution $\textless$ 10) were eliminated, targeting homogeneous regions such as water bodies, forests, or barren land. These filters yielded a refined dataset of 302,132 high-quality pairs, emphasizing urban centers and mixed-use landscapes.

CNSatMap distinguishes itself through its unprecedented scale, geographical diversity, and meticulous construction. Its high-resolution imagery, expansive coverage, and stringent quality controls establish it as a robust benchmark for algorithms bridging the visual-semantic gap between satellite imagery and cartographic representations. Potential applications span urban scene parsing, automated cartography, and geospatial feature extraction.

\section{Methodology}
\label{section:methods}

\subsection{Preliminaries}


Contemporary visual generation frameworks predominantly employ autoregressive modeling with next-token prediction objectives. In conventional AR paradigms, images undergo spatial compression through visual autoencoders to derive latent representations, which are subsequently quantized into discrete token sequences $\mathbf{x}=\left(x_1, x_2, \ldots, x_T\right)$. The AR model then sequentially predicts each token conditioned on its historical context:

\begin{equation}
	p(\mathbf{x}|c)=\prod_{t=1}^T p(x_t|x_{<t}, c)
\end{equation}where $c$ denotes optional conditioning signals. However, this token-wise generation paradigm manifests three fundamental limitations: (a) structural disintegration arising from the inherent conflict between unidirectional sequential modeling and the bidirectional spatial dependencies in visual data, (b) computational inefficiency stemming from obligatory raster-scan generation that prohibits parallel computation within spatial scales, and (c) constrained contextual reasoning capacity due to the absence of explicit hierarchical structural modeling in conventional AR frameworks.



To overcome these fundamental limitations, recent advances in autoregressive visual modeling have introduced a paradigm shift from token-level to scale-level prediction. The proposed next-scale prediction framework establishes hierarchical autoregression through three pivotal components: First, a multi-scale visual encoder (VQ-VAE\cite{van2017neural}) decomposes the input image $I$ into $K$ hierarchical feature maps $\left\{f_k\right\}_{k=1}^K$ with progressively increasing spatial resolution:

\begin{equation}
	f_k = \mathcal{E}_k(I) - \sum_{m=1}^{k-1} \mathcal{U}(\mathcal{Q}(f_m))
\end{equation}where $\mathcal{E}_k: \mathbb{R}^{H \times W \times 3} \rightarrow \mathbb{R}^{h_k \times w_k \times d}$ denotes the $k$-th scale encoder with spatial compression ratio $\left(H / h_k, W / w_k\right)$, $\mathcal{U}(\cdot)$ represents the spatial upsampling operator, and $\mathcal{Q}(\cdot)$ performs vector quantization. This residual formulation preserves spatial coherence while enabling explicit hierarchical decomposition.


Each continuous feature map $f_k \in \mathbb{R}^{h_k \times w_k \times d}$ undergoes codebook projection to discrete token maps through nearest-neighbor quantization:
\begin{equation}
	r_k^{(i,j)} = \underset{v\in\mathcal{V}}{\arg\min} |\mathcal{V}[v] - f_k^{(i,j)}|_2
\end{equation}where $\mathcal{V} \in \mathbb{R}^{|\mathcal{V}| \times d}$ represents the learned codebook and $r_k \in \mathcal{V}^{h_k \times w_k}$ constitutes the discrete representation at scale $k$.

The hierarchical autoregressive process then operates through scale-conditioned generation:
\begin{equation}
	p(\mathbf{r}|c) = \prod_{k=1}^K p(r_k|r_{<k}, c)
\end{equation}where $\mathbf{r}=\left(r_1, \ldots, r_K\right)$ represents the complete multi-scale representation. This formulation enables simultaneous prediction of all spatial positions within each scale while maintaining inter-scale dependencies, effectively resolving the limitations of conventional token-wise autoregression.

\subsection{Overview of EarthMapper}

EarthMapper introduces an advanced autoregressive framework tailored for controllable bidirectional translation between satellite imagery and cartographic maps, adeptly addressing the complexities of cross-modal geospatial synthesis. As depicted in Fig.~\ref{flowchart}, its core architecture hinges on the geo-conditioned joint scale autoregression (GJSA) mechanism, which orchestrates multi-scale feature alignment between input and output modalities within a unified training paradigm, facilitating seamless bidirectional translation. To enhance semantic fidelity, EarthMapper integrates a semantic infusion (SI) mechanism during training, minimizing feature-space disparities between synthetic and real images. During inference, a key point adaptive guidance (KPAG) mechanism dynamically modulates generation, balancing diversity and precision by anchoring outputs to salient geospatial features and adjusting conditioning strength based on image complexity.

\subsection{Hierarchical Residual Quantization}


For a given satellite image $\boldsymbol{S}$ and the corresponding map $\boldsymbol{M}$, an encoder $\boldsymbol{E}$ first transforms these inputs into latent feature maps $\boldsymbol{E}(\boldsymbol{S}), \boldsymbol{E}(\boldsymbol{M}) \in \mathbb{R}^{h \times w \times d}$, where $h \times w$ denotes the spatial dimensions and $d$ represents the feature dimensionality. To convert these continuous representations into discrete tokens suitable for autoregressive modeling, a baseline quantization step employs a learned codebook $\mathcal{Z}=\left\{z_k\right\}_{k=1}^K \subset \mathbb{R}^d$, where each $z_k$ is a prototype vector in the latent space. The quantization operation assigns each spatial feature vector $\hat{z}_{i j} \in \mathbb{R}^d$—extracted from $\boldsymbol{E}(\boldsymbol{S})$ or $\boldsymbol{E}(\boldsymbol{M})$—to its nearest codebook entry: 

\begin{equation}
z_q^{(i, j)}=\arg \min _{z_k \in \mathcal{Z}}\left\|\hat{z}_{i, j}-z_k\right\|_2
\end{equation}
where $z_q^{(i, j)}$ is the index of the selected codebook vector, and $\mathcal{Z}[z_q^{(i, j)}] \in \mathbb{R}^d$ denotes the corresponding discrete representation. This process establishes a foundation for token-based generation but struggles to capture the multi-scale intricacies inherent in high-resolution remote sensing imagery without incurring significant computational costs through an expanded codebook.

%

To address this limitation, inspired by \cite{tian2024visual}, we introduce a hierarchical residual quantization (HRQ) strategy, designed to efficiently encode the rich, multi-scale structure of geospatial data in a coarse-to-fine manner. Unlike traditional VQ, which relies on a single quantization step, HRQ employs a recursive residual approach with a fixed-size codebook $\mathcal{C} \subset \mathbb{R}^d$. For a given feature vector $z=\hat{z}_{i, j}$,HRQ generates a sequence of quantization indices:

\begin{equation}
\operatorname{HRQ}(z ; \mathcal{C}, D)=\left(k_1, k_2, \ldots, k_D\right)
\end{equation}where $k_d=\arg \min _{z_i \in \mathcal{C}}\left\|r_{d-1}-z_i\right\|_2$, with the residuals defined iteratively as $r_0=z$ and $r_d = r_{d-1}-\mathcal{C}\left[k_d\right] \text { for } d=1, \ldots, D .$ Here, $D$ denotes the quantization depth, controlling the granularity of the approximation. The reconstructed feature is approximated as $\hat{z} \approx \sum_{d=1}^D \mathcal{C}\left[k_d\right]$, enabling progressive refinement of residual errors. This approach compacts high-resolution representations into a hierarchical sequence of discrete indices, preserving fine-grained details and global structures without the computational overhead of large codebooks typical in conventional VQ-VAE frameworks.

Leveraging this HRQ mechanism, EarthMapper employs a multi-scale quantization autoencoder to discretize $\boldsymbol{S}$ and $\boldsymbol{M}$ into sets of token maps $\left\{\boldsymbol{s}_1, \boldsymbol{s}_2, \ldots, \boldsymbol{s}_k\right\}$ and $\left\{\boldsymbol{m}_1, \boldsymbol{m}_2, \ldots, \boldsymbol{m}_k\right\}$, respectively. At each scale $k$, the encoder $\boldsymbol{E}_k$ produces feature maps $\boldsymbol{E}_k(\boldsymbol{S}), \boldsymbol{E}_k(\boldsymbol{M}) \in \mathbb{R}^{h_k \times w_k \times d}$, with spatially varying resolutions $h_k \times w_k$ reflecting the hierarchical decomposition. These are then quantized via HRQ:

\begin{equation}
\begin{aligned}
	\mathbf{s}_k^{(i, j)} & =\operatorname{HRQ}\left(\boldsymbol{E}_k(\mathbf{S})^{(i, j)} ; \mathcal{C}, D\right), \\
	\mathbf{m}_k^{(i, j)} & =\operatorname{HRQ}\left(\boldsymbol{E}_k(\mathbf{M})^{(i, j)} ; \mathcal{C}, D\right),
\end{aligned}
\end{equation}yielding token maps $\boldsymbol{s}_k, \boldsymbol{m}_k \in \mathbb{Z}^{h_k \times w_k \times D}$ that encode multi-depth indices at each spatial position. This multi-scale tokenization captures the hierarchical nature of remote sensing data, aligning with the perceptual progression from coarse (e.g., regional topography) to fine (e.g., infrastructure details) scales. By integrating spatial coherence and resolution adaptability, HRQ ensures that EarthMapper meets the stringent fidelity requirements of BSMT task, enhancing both the realism and utility of the generated outputs for applications such as urban planning and environmental monitoring.

\subsection{Geo-Conditioned Joint Scale Autoregression}

In EarthMapper, we propose a pioneering geo-conditioned autoregressive framework to jointly model satellite imagery and corresponding maps, leveraging geographic coordinates as a spatially-informed control signal. Unlike prior works such as ControlVAR~\cite{li2024controlvar}, which primarily explore class-level controls for general imagery, our approach introduces a domain-specific innovation by conditioning the AR process on geospatial context. This enables the generation of satellite image-map pairs that are inherently aligned with real-world locations, addressing a critical need in remote sensing applications.

We first reformulate geographic coordinates $g=(\phi, \lambda)$ as a continuous control signal through sinusoidal positional encoding: 

\begin{equation}
\boldsymbol{c}_g=\operatorname{MLP}(\operatorname{Sin} \operatorname{Embed}(\phi) \oplus \operatorname{SinEmbed}(\lambda))
\end{equation}where $\phi$ and $\lambda$ denote latitude and longitude, respectively, and $\oplus$ represents concatenation. This encoding scheme maintains spatial continuity across the Earth's surface, effectively accounting for the non-Euclidean properties inherent in geographic coordinates.

After that, We redefine the conditional AR generation task to model the joint distribution $p(\boldsymbol{S}, \boldsymbol{M} \mid c_g)$. Building on the multi-scale tokenization established by HRQ in the previous subsection, we derive discrete multi-scale feature maps for satellite imagery and maps, represented as:

\begin{equation}
\left\{\boldsymbol{s}_k\right\}_{k=1}^K=\Phi_S(\boldsymbol{S}), \quad\left\{\boldsymbol{m}_k\right\}_{k=1}^K=\Phi_M(\boldsymbol{M})
\end{equation}where both modalities are decomposed into $K$ hierarchical scales using shared codebooks, this approach ensures geometric consistency between the latent spaces of satellite imagery and map representations across all scales.


To preserve the autoregressive property while capturing cross-modal dependencies, we pair tokens at each scale into a joint representation $\boldsymbol{r}_k=\left(\boldsymbol{s}_k, \boldsymbol{m}_k\right)$ and define the generative process as:

\begin{equation}
p(\boldsymbol{S}, \boldsymbol{M} \mid c_g)=\prod_{k=1}^K p\left(\boldsymbol{r}_k \mid \boldsymbol{r}_{<k}, c_g\right)
\end{equation}where $\boldsymbol{r}_{<k}=\left\{\boldsymbol{r}_1, \boldsymbol{r}_2, \ldots, \boldsymbol{r}_{k-1}\right\}$ denotes the sequence of prior token pairs. This hierarchical joint modeling is particularly crucial for the BSMT task, where the geometric alignment between satellite features (e.g., building footprints) and their map representations (e.g., polygon symbols) must be preserved across scales.

%
%
%
%

Following \cite{tian2024visual}, EarthMapper employs a GPT-2-style Transformer architecture to implement the geo-conditioned AR process. Joint modeling optimizes the joint likelihood using a cross-entropy loss, supervising the prediction of $\boldsymbol{r}_{t}$ given $\boldsymbol{r}_{<t}$ and $c_g$:
\begin{equation}
	\mathcal{L}_{\mathrm{jot}}=\mathbb{E}_{\boldsymbol{r} \sim p(\boldsymbol{r} \mid c_g)}\left[-\sum_{k=1}^K \log p\left(\boldsymbol{r}_k \mid \boldsymbol{r}_{<k}, c_g\right)\right]
\end{equation}where $\mathcal{L}_{\mathrm{jot}}$ presents the joint modeling loss. 

\subsection{Semantic Infusion for Geovisual Coherence}

Although the geo-conditioned AR framework in EarthMapper adeptly generates structurally aligned satellite imagery and maps based on geographic coordinates, ensuring geovisual coherence—where outputs exhibit semantically meaningful and contextually consistent features—poses a challenge. To address this, we introduce a semantic infusion mechanism that integrates a pre-trained visual foundation model into the framework, leveraging its rich feature space to enrich the AR model's latent representations with general geovisual knowledge, thereby enhancing the perceptual realism and utility of the generated satellite-map pairs for BSMT tasks.

Let $\mathcal{G}(\cdot)$ denote the Transformer-based AR model from the previous subsection, which predicts token pairs $\boldsymbol{r}_{k} = \left(\boldsymbol{s}_k, \boldsymbol{m}_k\right)$ given prior tokens $\boldsymbol{r}_{<k}$ and the geo-control signal $c_g$. The pre-trained vision model's encoder $\mathcal{E}_{\mathrm{sem}}(\cdot)$ extracts semantic features from the input satellite imagery $\boldsymbol{S}$ and maps $\boldsymbol{M}$. Then we introduce a lightweight alignment network $\mathcal{A}$ to bridge the dimensionality gap between the AR model's embedding space and the feature space of a pre-trained visual foundation model: 

\begin{equation}
\mathcal{A}\left(\mathcal{G}\left(\boldsymbol{r}_{<k}, c_g\right)\right) \in \mathbb{R}^{h_k \times w_k \times d_{\mathrm{sem}}},
\end{equation} where $h_k \times w_k$ corresponds to the spatial resolution at scale $k$, $d_{\mathrm{sem}}$ is the dimensionality of the semantic space. This aligned representation is then constrained to reflect the semantic context provided by $\mathcal{E}_{\mathrm{sem}}(\cdot)$.

To guide the AR model toward geovisual coherence, we introduce a semantic infusion loss $\mathcal{L}_{\text {sem }}$ that minimizes the discrepancy between the aligned AR embeddings and the pre-trained foundation model's semantic features. For a given scale $t$, we define:

\begin{equation}
\mathcal{L}_{\mathrm{sem}}=\frac{1}{K} \sum_{k=1}^K\left\|\mathcal{A}\left(\mathcal{F}\left(\boldsymbol{r}_{<k}, c_g\right)\right)-\mathcal{E}_{\mathrm{sem}}\left(\boldsymbol{r}_k\right)\right\|_2^2
\end{equation}where $\mathcal{E}_{\mathrm{sem}}\left(\boldsymbol{r}_k\right)$ approximates the semantic features of the token pair $\boldsymbol{r}_k$ by mapping the decoded representations $\Phi_S^{-1}\left(s_k\right)$ and $\Phi_M^{-1}\left(\boldsymbol{m}_k\right)$—reconstructed from the HRQ tokenizer—back into the semantic space. This loss ensures that the AR predictions align with high-level geospatial semantics without overriding the joint modeling objective.

The overall training objective combines the joint modeling loss $\mathcal{L}_{\mathrm{jot}}$ from the geo-conditioned AR with the semantic infusion loss $\mathcal{L}_{\mathrm{sem}}$, weighted by a hyperparameter $\sigma$:

\begin{equation}
\mathcal{L}=\mathcal{L}_{\mathrm{jot}}+\sigma \cdot \mathcal{L}_{\mathrm{sem}},
\end{equation}During training, the foundation model $\mathcal{E}_{\mathrm{sem}}$ remains frozen, while the parameters of $\mathcal{G}$ and $\mathcal{A}$ are optimized jointly.

%
%
%

%

\begin{figure}[tbp]
	\begin{center}
		\centerline{\includegraphics[width=1\linewidth]{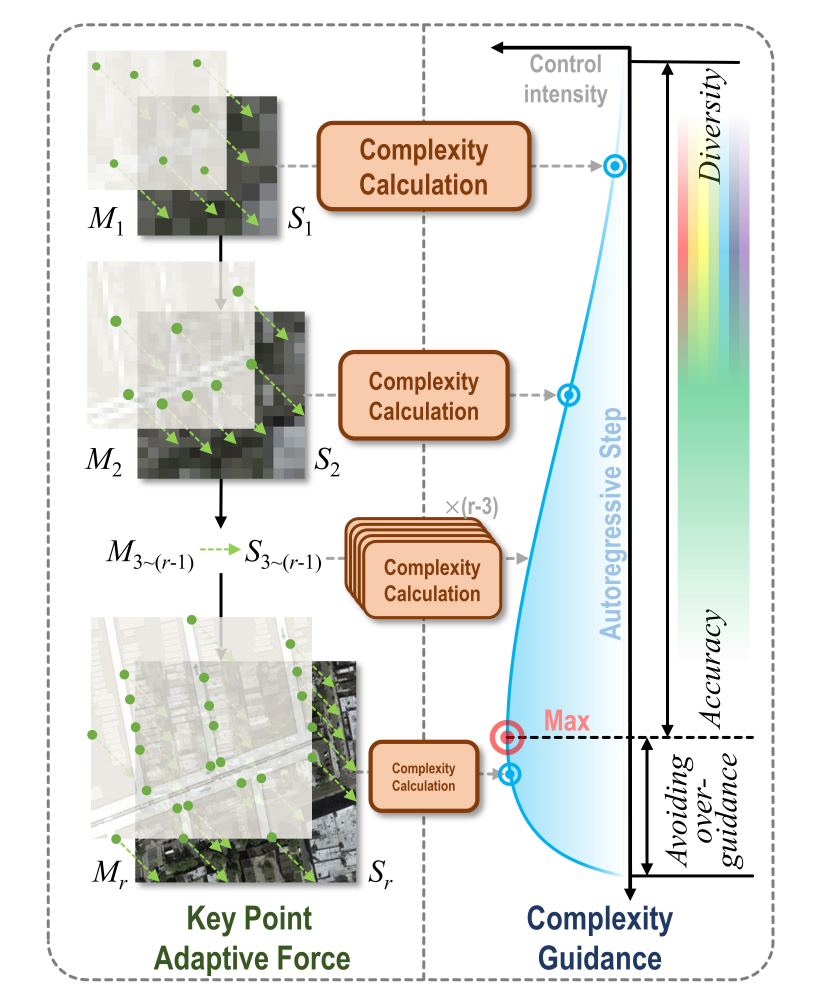}}
		\caption{Schematic diagram of the inference section, with keypoint adaptive forcing on the left and complexity bootstrapping on the right, both of which are connected by the computation of complexity and together comprise the inference.}\label{KAG}
		\vspace{-10pt} 
	\end{center}
\end{figure}

\subsection{Key Point Force}
\label{section:submethod1}

During the training phase of EarthMapper, the model is designed to learn to generate corresponding satellite image and map pairs $[	\mathbf{s},	\mathbf{m}]$ based on input geographic coordinates $g$. In the inference phase, the task of generating satellite images from maps is taken as an example: the map acts as the conditional image, while the satellite image serves as the generation target (the reverse task works on a similar principle). Specifically, In the first autoregressive step, this geographic coordinate vector $c_g$ is employed to predict $[\mathbf{s},	\mathbf{m}]$. However, geographic coordinates $g$ alone provide only coarse-grained information and lack the precision needed to capture fine details within the image. To enhance the precision of the control, the map portion of the generated paired image is substituted with a vector-quantized real map:
\begin{equation}
	Q_{\mathbf{m}} = \operatorname{HRQ}((\boldsymbol{E}(\mathbf{m}))
\end{equation}where $Q_{\mathbf{m}}$ denotes the map tokens after vector quantization, $ \boldsymbol{E}(\mathbf{m}) \in \mathbb{R}^{h \times w \times d}$, where $ \boldsymbol{E}$ denotes the visual
encoder, $h \times w$ denotes the spatial dimensions and $d$ represents the feature dimensionality. thereby imposing additional constraints on the model's generation output.

The architectural design of EarthMapper precludes direct information exchange between the generated paired images at each step, thus rendering the effectiveness of the conditional control contingent on the optimisation achieved during training. To address this limitation, we propose the key point force method, as shown on the left side of Fig.~\ref{KAG}. Let $C$ denote the conditional image and $G$ the generated image. After performing vector quantization on $C$, we obtain the quantized indices $Q_C = \{ q_{C,1}, q_{C,2}, \ldots, q_{C,N} \}$, where each $q_{C,i} \in \{0,1,\ldots,K-1\}$ and $K$ is the number of codebook vectors. We normalize these indices to the interval $[0,1]$ as follows:
\begin{equation}
	\hat{q}_{C,i} = \frac{q_{C,i}}{K-1}, \quad \forall i = 1,2,\ldots,N
\end{equation}A threshold $\tau$ is established, and the set of key points is defined as:
\begin{equation}
	\mathcal{K} = \{ i \mid \hat{q}_{C,i} > \tau \}
\end{equation}During the autoregressive generation of $G$, for each position $i = 1$ to $N$, we sample the index $q_{G,i}$ from the conditional distribution:
\begin{equation}
	q_{G,i} \sim p(q_{G,i} | q_{G,1}^{\text{final}}, q_{G,2}^{\text{final}}, \ldots, q_{G,i-1}^{\text{final}}, C)
\end{equation}Then, the final index for position $i$ is set as:
\begin{equation}
	q_{G,i}^{\text{final}} = \begin{cases} 
		q_{G,i} & \text{if } i \notin \mathcal{K} \\
		\max(0, \min(q_{G,i} + q_{C,i}, K-1)) & \text{if } i \in \mathcal{K} 
	\end{cases}
\end{equation}This integrates the information from the key points into the generated image at each stage of the autoregressive process.

This approach involves the use of a small set of key points to restrict the generation process, thereby ensuring that conditional information has a pronounced effect on the generated image. Additionally, the implementation of a key point selection mechanism, in conjunction with index addition at the vector quantization level, ensures the provision of sufficient input information, while concurrently avoiding the excessive control that could potentially suppress image diversity. This design is intended to preserve a certain degree of variability in the generated images, thereby aligning them more closely with the characteristics of real images than merely replicating the style of the conditional image.

\subsection{Complexity Guidance}

In the existing conditional image generation inference processes, such as diffusion models, a commonly used method is to balance the condition weights and the model's autonomous generation weights through a classifier-free guidance (CFG) to enhance the quality of image generation. In the context of autoregressive image generation models, the CFG method can be formulated as follows:
\begin{align}
	p(x_i|x_{<i}, c) 
	&= p(x_i|x_{<i}, \emptyset) \notag \\
	&\quad + s \cdot \left(p(x_i|x_{<i}, c) - p(x_i|x_{<i}, \emptyset)\right)
\end{align}where $p(x_i|x_{<i})$ denotes the probability distribution of the current pixel $x_i$ conditioned on the previously generated pixels $x_{<i}$. The unconditional distribution $p(x_i|x_{<i}, \emptyset)$ represents the model's autonomous generation capability without any conditional guidance, while the conditional distribution $p(x_i|x_{<i}, c)$ represents the model's generation capability guided by the condition $c$. The CFG method balances these two distributions using a guidance strength parameter $s$.

However, for EarthMapper's multi-resolution autoregressive image generation structure, a single fixed CFG guidance strength cannot fully exploit the potential of the autoregressive generation model. To address this, a novel conditional guidance method is proposed, termed complexity guidance (CG), as shown on the right side of Fig.~\ref{KAG}, which introduces a dynamic guidance strength $s(x_i, \phi)$ during the sampling process.

Let $r_i$ denote the resolution level at generation step $i$, and $C(x_i)$ represent the complexity of the generated image $x_i$. We introduce two dynamic parameters: $\alpha(r_i)$ and $\beta(C(x_i))$, which are functions of the resolution level and image complexity, respectively. The dynamic guidance strength $s(x_i, \phi)$ can be formulated as:
\begin{equation} s(x_i, \phi) = \gamma \cdot \alpha(r_i) \cdot \beta(C(x_i)) \end{equation}
where $\gamma$ is a hyperparameter to control the overall strength of the complexity-guided CFG.

In the initial phases of image generation, characterized by low resolution, the model employs rudimentary coordinate vectors to predict, thereby yielding images of minimal complexity. At this stage, it is necessary to reduce the condition guidance strength to enhance the diversity of the generation results and to prevent the model from prematurely falling into erroneous generation paths. This can be achieved by setting $\alpha(r_i)$ to a low value when $r_i$ is small. As the generation process advances and the resolution increases, $\alpha(r_i)$ should progressively increase to ensure the accuracy and precision of the generated images.

Concurrently, the autoregressive image generation process entails a dynamic adjustment from the addition of details to the removal of redundant details. This process can be quantified through image complexity $C(x_i)$. In the initial phase of generation, complexity is minimal, necessitating only a modest amount of condition guidance to facilitate the extensive addition of details. As the generation process progresses, the complexity increases, reaching a peak as a result of the incorporation of a substantial number of details. At this stage, it becomes essential to augment the guidance strength by setting $\beta(C(x_i))$ to a high value. In the subsequent phase, the removal of redundant details leads to a modest decline in complexity, necessitating a corresponding reduction in $\beta(C(x_i))$ to avert the retention of invalid details resulting from excessive constraint.

The complexity $C(x_i)$ can be quantified using various measures, such as the entropy of the image or the number of distinct features present. One possible formulation is:
\begin{equation} C(x_i) = -\sum_{j} p(x_{i,j}) \log p(x_{i,j}) \end{equation}
where $x_{i,j}$ represents the $j$-th pixel or feature of the generated image $x_i$, and $p(x_{i,j})$ is the probability distribution over the pixels or features.

By incorporating the dynamic guidance strength $s(x_i, \phi)$ into the CFG formulation, we obtain the complexity-guided conditional function (CG-CF) for autoregressive image generation:
\begin{align}
	p_{\theta}(x_i|x_{<i}, c) &= p_{\theta}(x_i|x_{<i}) \notag \\
	&\quad + s(x_i, \phi) \cdot \nabla_{p_{\theta}(x_i|x_{<i})} \log p_{\varphi}(c|x_i)
\end{align}
where $p_{\theta}(x_i|x_{<i}, c)$ represents the conditional probability distribution, $p_{\theta}(x_i|x_{<i})$ represents the unconditional probability distribution, $s(x_i, \phi)$ is the dynamic guidance strength , and $\nabla_{p_{\theta}(x_i|x_{<i})} \log p_{\varphi}(c|x_i)$ represents the gradient of the conditional probability $p_{\varphi}(c|x_i)$ with respect to the unconditional probability $p_{\theta}(x_i|x_{<i})$.

This complexity guidance method enables the model to adaptively balance between its autonomous generation capability and the conditional guidance based on the resolution level and complexity of the current generation context, thereby improving the realism and precision of the generated images in EarthMapper's bidirectional translation tasks between satellite images and maps.

\section{Experiments}
\label{section:EXPERIMENTS}


\subsection{Dataset and Evaluation Metrics}

We evaluate our approach on two datasets: the New York dataset\cite{isola2017image} and our constructed CNSatMap dataset.

The New York dataset consists of paired aerial photographs and corresponding maps obtained from Google Maps, primarily covering New York City and its surrounding areas. It includes 1,096 training pairs, 1,098 validation pairs, and 1,098 test pairs, each at a resolution of 600$\times$600 pixels. The dataset is geographically partitioned along latitudinal lines with a buffer zone to eliminate spatial overlap, establishing a robust benchmark for BSMT tasks.

The CNSatMap dataset comprises 302,132 georeferenced image pairs for cross-modal translation between satellite imagery and cartographic maps. It integrates high-resolution satellite images (0.6 m/pixel) from Google Earth and vector-based maps from Tianditu, spanning 38 major Chinese cities. Images are tiled into 256$\times$256 pixels in the EPSG:3857 projection and undergo rigorous quality filtering to exclude cloud-covered, oblique, or homogeneous regions. The dataset is divided into 185,751 training, 58,190 validation, and 58,191 test pairs, serving as a large-scale and high-quality benchmark for urban scene understanding and geospatial image translation.

For the map-to-satellite translation task, an open-domain generation problem, we employ Fr\text {é}chet inception distance (FID)\cite{heusel2017gans}, kernel inception distance (KID)\cite{binkowski2018demystifying}, Precision, and Recall\cite{kynkaanniemi2019improved} to evaluate visual realism, semantic plausibility, and distribution consistency of the generated images. For the satellite-to-map translation task, a structured reconstruction problem, we utilize mean square error (MSE), peak signal-to-noise ratio (PSNR), structural similarity index (SSIM), and Learned Perceptual Image Patch Similarity (LPIPS)\cite{zhang2018unreasonable} to assess pixel-level accuracy, structural preservation, and perceptual similarity, respectively.

\subsection{Experimental Setup}

All images are resized to a resolution of 256$\times$256 during both training and inference. We adopt a GPT-2-style transformer architecture with a depth of 24 layers, initialized with pretrained weights from the VAR model to accelerate convergence. For semantic feature fusion, the pretrained DINOv2\cite{oquab2023dinov2} model is incorporated as the visual backbone. The balancing hyperparameter $\sigma$ is set to 0.5. Training is conducted on eight NVIDIA A800 GPUs, each with 80 GB of memory, using the AdamW\cite{loshchilov2017decoupled} optimizer. The model is trained for 100 epochs with a batch size of 24, ensuring a trade-off between computational efficiency and gradient stability. During inference, we employ top-$k$ and top-$p$ sampling strategies with $k$=100 and $p$=0.55 to enhance the quality and diversity of the generated outputs.

%
%
%
%
%
%
%
%

\begin{table*}[htbp]
	\centering
	\caption{Performance Comparison on New York Dataset. Optimal values are highlighted in \textcolor{red}{\textbf{bold red}} and sub-optimal values in \textcolor{blue}{\textbf{bold blue}}.}
	\label{NY_comparison}
	\renewcommand\arraystretch{1.0}
	\setlength{\tabcolsep}{6pt}
	\footnotesize
	\resizebox{\textwidth}{!}{
		\begin{tabular}{@{}c|l|cccc|cccc@{}}
			\Xhline{2\arrayrulewidth}
			\multirow{2}{*}{Category} & \multirow{2}{*}{Method} & \multicolumn{4}{c|}{Map-to-Satellite Translation} & \multicolumn{4}{c}{Satellite-to-Map Translation} \\ 
			\cline{3-10}
			& & FID $\downarrow$ & KID $\downarrow$ & Precision $\uparrow$ & Recall $\uparrow$ & SSIM $\uparrow$ & PSNR $\uparrow$ & RMSE $\downarrow$ & LPIPS $\downarrow$ \\ 
			\hline
			
			\multirow{3}{*}{GANs} 
			& CycleGAN & 104.96 & 4.93 & 0.0580 & \textcolor{blue}{\textbf{0.3439}} & 0.6344 & 23.55 & \textcolor{blue}{\textbf{0.0554}} & 0.4345  \\
			& Pix2Pix & 86.36 & 4.10 & 0.0334 & 0.2986 & 0.5892 & 20.61 & 0.1460 & 0.4697 \\
			& StegoGAN & 121.96 & 6.39 & 0.0928 & 0.2456 & 0.6422 & \textcolor{blue}{\textbf{24.70}} & \textcolor{red}{\textbf{0.0530}} & 0.4385  \\
			\hline
			\multirow{3}{*}{LDMs} 
			& BBDM & 196.51 & 14.79 & 0.0190 & 0.1087 & 0.6303 & 23.71 & 0.0692 & 0.3211 \\
			& ControlNet & 133.83 & 7.18 & \textcolor{blue}{\textbf{0.6017}} & 0.0229 & 0.6043 & 21.12 & 0.1058 & 0.4395 \\
			& UniControl & 101.05 & 4.27 & 0.3193 & 0.1100 & 0.3842 & 11.49 & 0.3001 & 0.4242  \\
			
			\hline
			
			\multirow{3}{*}{ARs} 
			& CAR & 80.37 & 5.20 & 0.3749 & 0.1691 & 0.5831 & 23.57 & 0.0688 & 0.3602 \\
			& ControlVAR & \textcolor{blue}{\textbf{58.23}} & \textcolor{blue}{\textbf{2.37}} & 0.5008 & 0.2151 & \textcolor{blue}{\textbf{0.6465}} & 24.40 & 0.0647 & \textcolor{blue}{\textbf{0.3156}} \\
			& EarthMapper & \textcolor{red}{\textbf{36.54}} & \textcolor{red}{\textbf{0.99}} & \textcolor{red}{\textbf{0.6182}} & \textcolor{red}{\textbf{0.4890}} & \textcolor{red}{\textbf{0.6534}} & \textcolor{red}{\textbf{25.04}} & 0.0611 & \textcolor{red}{\textbf{0.2819}} \\
			
			\Xhline{2\arrayrulewidth}
		\end{tabular}
	}
	\vspace{0.2cm}
	\parbox{\textwidth}{\footnotesize \textit{Note:} Arrows indicate desired direction of improvement ($\downarrow$ lower is better, $\uparrow$ higher is better).}
\end{table*}

\begin{table*}[htbp]
	\centering
	\caption{Performance Comparison on CNSatMap Dataset. Optimal values are highlighted in \textcolor{red}{\textbf{bold red}} and sub-optimal values in \textcolor{blue}{\textbf{bold blue}}.}
	\label{CNSatMap_comparison}
	\renewcommand\arraystretch{1.0}
	\setlength{\tabcolsep}{6pt}
	\footnotesize
	\resizebox{\textwidth}{!}{
		\begin{tabular}{@{}c|l|cccc|cccc@{}}
			\Xhline{2\arrayrulewidth}
			\multirow{2}{*}{Category} & \multirow{2}{*}{Method} & \multicolumn{4}{c|}{Map-to-Satellite Translation} & \multicolumn{4}{c}{Satellite-to-Map Translation} \\ 
			\cline{3-10}
			& & FID $\downarrow$ & KID $\downarrow$ & Precision $\uparrow$ & Recall $\uparrow$ & SSIM $\uparrow$ & PSNR $\uparrow$ & RMSE $\downarrow$ & LPIPS $\downarrow$ \\ 
			\hline
			
			\multirow{3}{*}{GANs} 
			& CycleGAN & 163.96 & 11.04 & 0.0946 & 0.3646 & 0.6521 & 22.05 & 0.0819 & 0.3716 \\
			& Pix2Pix & 150.85 & 12.50 & 0.0879 &\textcolor{blue}{\textbf{0.4334}} & 0.6384 & 23.26 & 0.0701 & 0.3604 \\
			& StegoGAN & 73.55 & 4.89 & 0.1652 & 0.3722 & 0.6635 & 25.30 & 0.0577 & 0.3118 \\
			\hline
			\multirow{3}{*}{LDMs} 
			& BBDM & 110.98 & 9.48 & 0.0584 & 0.4110 & \textcolor{blue}{\textbf{0.7184}} & \textcolor{blue}{\textbf{26.36}} & \textcolor{blue}{\textbf{0.0519}} & \textcolor{red}{\textbf{0.2954}} \\
			& ControlNet &87.08 & 6.79 & \textcolor{red}{\textbf{0.4886}} & 0.2110 & 0.5685 & 19.20 & 0.1265 & 0.3947  \\
			& UniControl & 235.70 & 20.42 & 0.0372 & 0.4284 & 0.4945 & 17.78 & 0.1951 & 0.5753\\
			
			\hline
			
			\multirow{3}{*}{ARs} 
			& CAR &55.29 & 4.87 & 0.2216 & \textcolor{red}{\textbf{0.4634}} & 0.5584 & 22.15 & 0.0831 & 0.4631 \\
			& ControlVAR & \textcolor{blue}{\textbf{53.07}} &\textcolor{blue}{\textbf{3.49}} & 0.3134 & 0.3422 & 0.7114 & 25.29 & 0.0599 & 0.3542 \\
			& EarthMapper & \textcolor{red}{\textbf{29.89}} & \textcolor{red}{\textbf{2.06}} & \textcolor{blue}{\textbf{0.4294}} & 0.3954 & \textcolor{red}{\textbf{0.7300}} & \textcolor{red}{\textbf{26.88}} & \textcolor{red}{\textbf{0.0510}} & \textcolor{blue}{\textbf{0.3103}} \\
			
			\Xhline{2\arrayrulewidth}
		\end{tabular}
	}
	\vspace{0.2cm}
	\parbox{\textwidth}{\footnotesize \textit{Note:} Arrows indicate desired direction of improvement ($\downarrow$ lower is better, $\uparrow$ higher is better).}
\end{table*}

\begin{figure*}[tbp]
	\begin{center}
		\centerline{\includegraphics[width=1\linewidth]{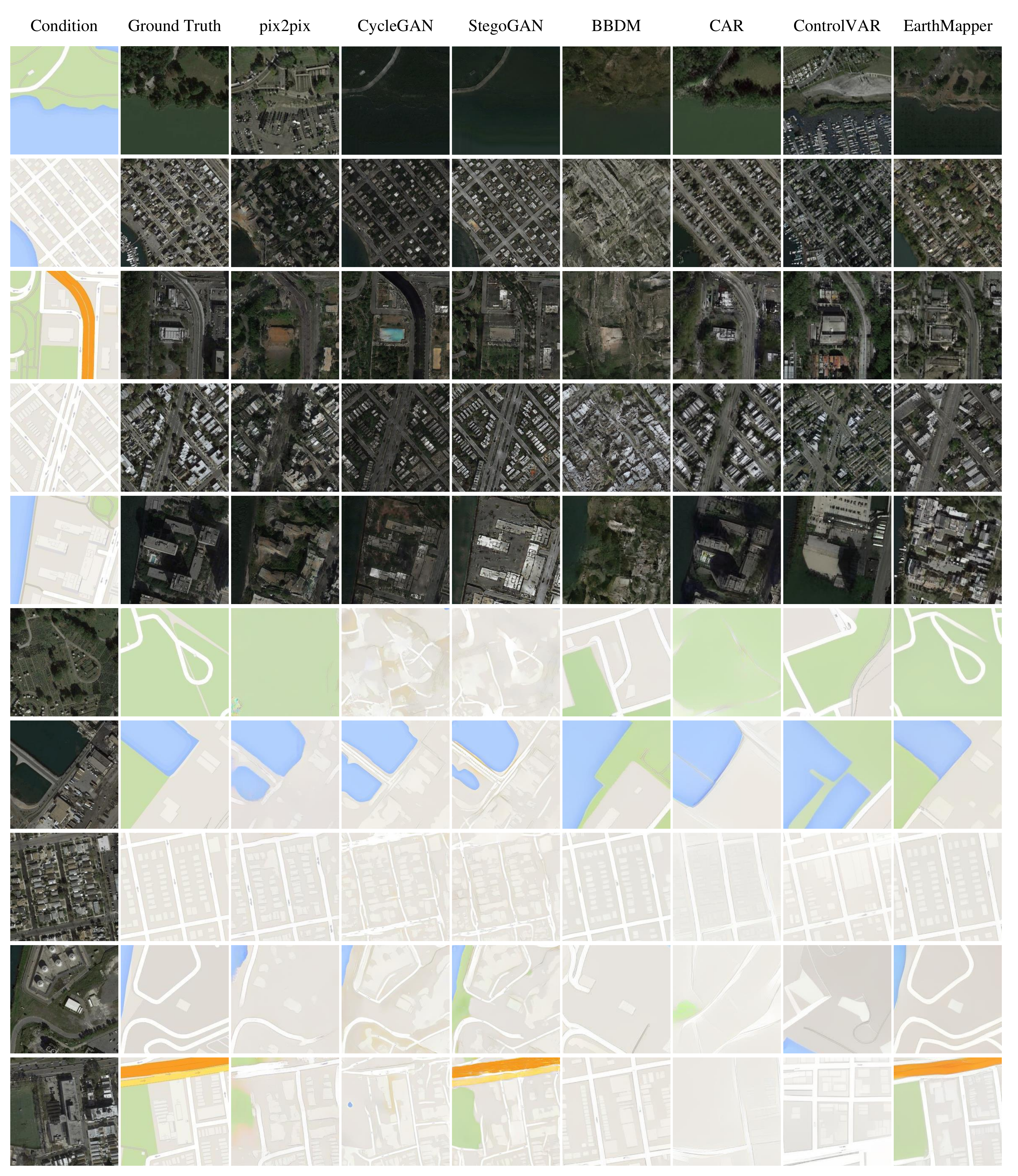}}
		\caption{Qualitative comparison of bidirectional satellite-map  translation results on the New York test set. The top five rows illustrate map-to-satellite translation, and the bottom five rows depict satellite-to-map translation.}\label{vis_NY}
	\end{center}
\end{figure*}

\begin{figure*}[tbp]
	\begin{center}
		\centerline{\includegraphics[width=1\linewidth]{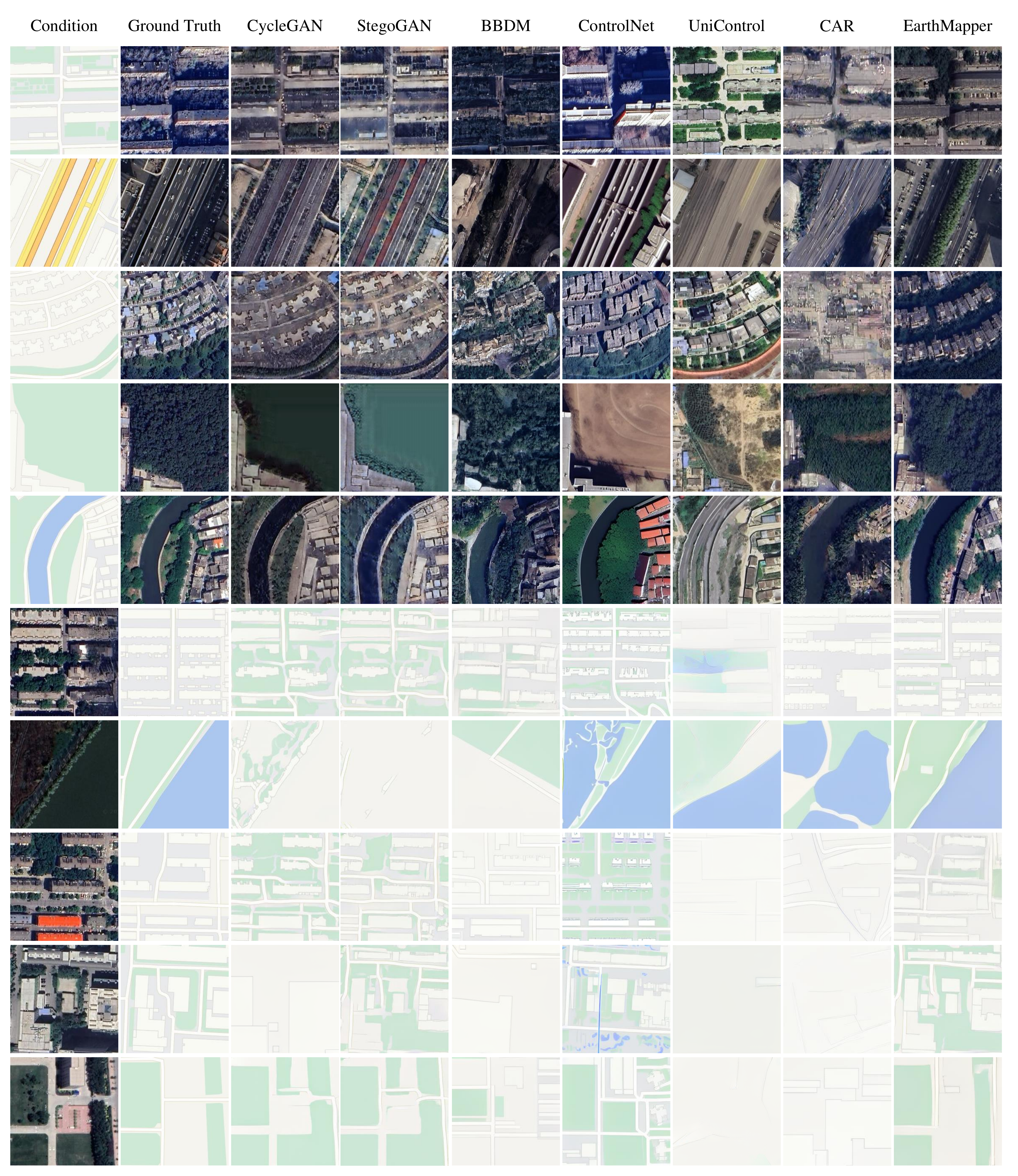}}
		\caption{Qualitative comparison of bidirectional satellite-map translation results on the CNSatMap test set. The top five rows illustrate map-to-satellite translation, and the bottom five rows depict satellite-to-map translation.}\label{vis_CNSatMap}
	\end{center}
\end{figure*}

\subsection{Comparison with State-of-the-art Methods}

To assess the robustness and generalizability of the proposed EarthMapper framework, we conducted a comparative evaluation on both the New York and CNSatMap datasets, with results presented in Tables~\ref{NY_comparison} and~\ref{CNSatMap_comparison}, respectively. These experiments encompass BSMT tasks, benchmarking EarthMapper against representative methods from GANs (CycleGAN\cite{zhu2017unpaired}, Pix2Pix\cite{isola2017image}, and StegoGAN\cite{wu2024stegogan}), LDMs (BBDM\cite{li2023bbdm}, ControlNet\cite{zhang2023adding}, UniControl\cite{qin2023unicontrol}), and ARs (CAR\cite{yao2024car}, ControlVAR\cite{li2024controlvar}). Performance is quantified using a suite of metrics—FID, KID, Precision, Recall, SSIM, PSNR, RMSE, and LPIPS—capturing generation quality, diversity, and structural fidelity. The optimal and suboptimal performances are distinctly highlighted in \textcolor{red}{\textbf{red}} and \textcolor{blue}{\textbf{blue}}, respectively.

In the map-to-satellite translation task, EarthMapper achieves an FID of 36.54 on New York and 29.89 on CNSatMap, significantly outperforming GAN-based methods like CycleGAN (104.96 and 163.96) and Pix2Pix (86.36 and 150.85). This substantial gap reflects EarthMapper’s ability to align the feature distributions of generated and real images closely, minimizing the Fréchet distance and enhancing visual realism—a critical advantage over GANs, which exhibit higher distributional mismatch. Similarly, its KID of 0.99 (New York) and 2.06 (CNSatMap) surpasses ControlVAR’s 2.37 and 3.49, indicating finer semantic consistency through reduced kernel-based discrepancies, unlike LDMs such as BBDM (14.79 and 9.48), which struggle with coarser feature alignment.

EarthMapper’s Precision of 0.6182 on New York and 0.4294 on CNSatMap demonstrates high fidelity, with generated samples effectively residing within the real image manifold. Although ControlNet achieves a higher Precision of 0.4886 on CNSatMap, EarthMapper’s Recall of 0.4890 (New York) and 0.3954 (CNSatMap) exceeds ControlNet’s 0.2110, showcasing superior diversity and coverage of the real distribution. Compared to GANs like StegoGAN (Recall: 0.2456 on New York, 0.3722 on CNSatMap), EarthMapper balances quality and diversity more effectively, avoiding the overfitting tendencies seen in CycleGAN’s inflated Recall (0.3439 and 0.3646). This equilibrium underscores EarthMapper’s autoregressive strength in modeling complex, open-domain dependencies, delivering realistic and varied satellite imagery.

For the satellite-to-map translation task, EarthMapper records an SSIM of 0.6534 on New York and 0.7300 on CNSatMap, outperforming ControlVAR (0.6465 and 0.7114) and BBDM (0.6303 and 0.7184). This indicates superior preservation of luminance, contrast, and structural details, critical for map reconstruction, where EarthMapper minimizes covariance discrepancies more effectively than GANs like CycleGAN (0.6344 and 0.6521). Its PSNR of 25.04 (New York) and 26.88 (CNSatMap) exceeds StegoGAN’s 24.70 and BBDM’s 26.36, reflecting lower pixel-level error and higher signal fidelity, a testament to its precise reconstruction capabilities over LDMs like UniControl (11.49 on New York).

\begin{figure}[tbp]
	\begin{center}
		\centerline{\includegraphics[width=1\linewidth]{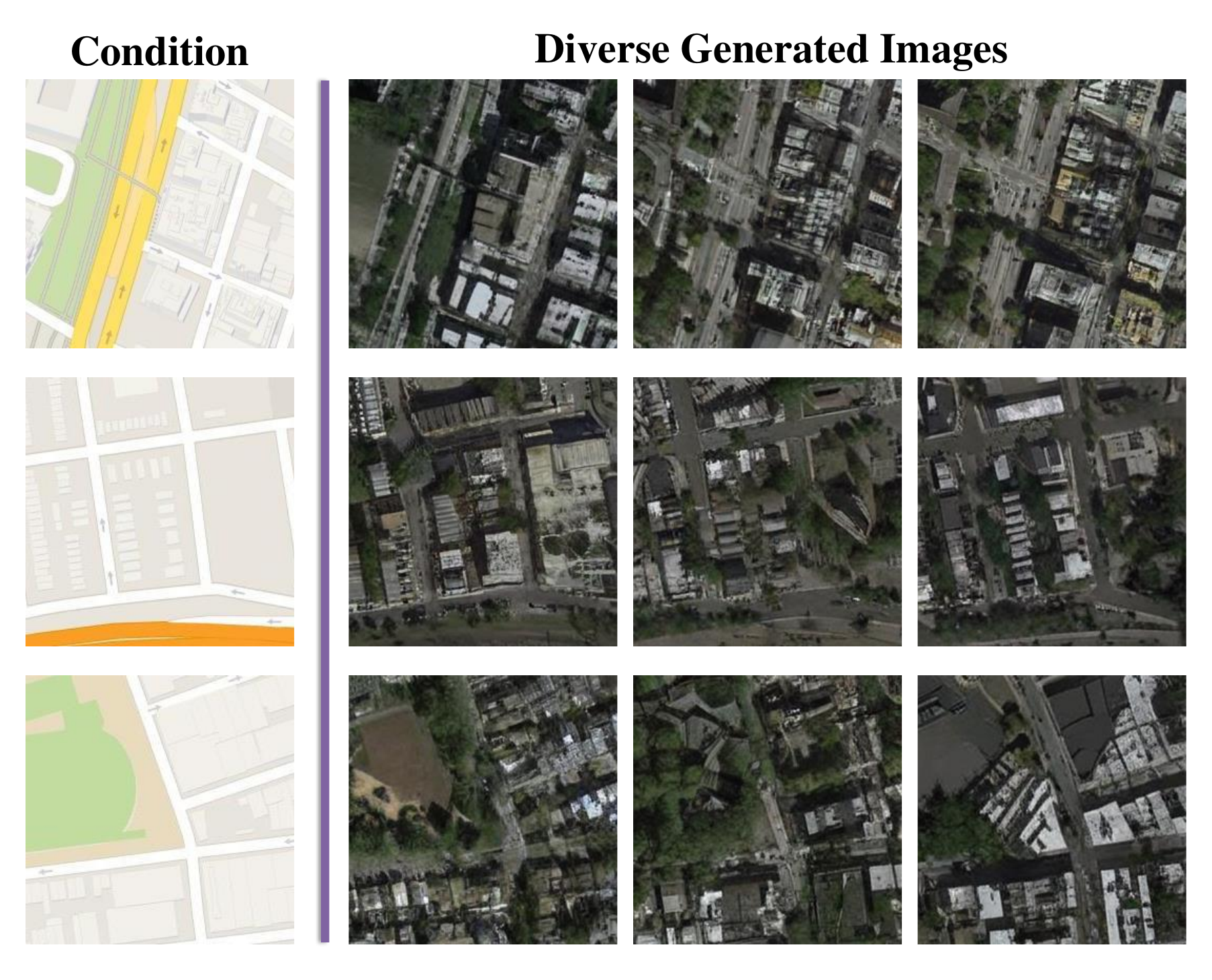}}
		
		\caption{Illustration of generative diversity in EarthMapper. Given a single conditional input (left), our method produces multiple distinct outputs (right), showcasing varied yet semantically consistent satellite and map representations. }\label{diverse}
	\end{center}
\end{figure}

EarthMapper’s RMSE of 0.0611 (New York) and 0.0510 (CNSatMap) edges out ControlVAR (0.0647 and 0.0599), achieving pixel accuracy comparable to StegoGAN’s 0.0530 (New York) and BBDM’s 0.0519 (CNSatMap). This narrow margin highlights its robustness in minimizing pixel discrepancies, crucial for structured outputs. In terms of LPIPS, EarthMapper’s 0.2819 (New York) and 0.3103 (CNSatMap) closely rival BBDM’s 0.2954 (CNSatMap), indicating strong perceptual similarity despite nuanced competition from LDMs. Unlike ControlNet (0.4395 on New York, 0.3947 on CNSatMap), EarthMapper maintains consistency across datasets, avoiding the perceptual degradation seen in GANs like Pix2Pix (0.4697 and 0.3604).

%

The visualization results in Fig.\ref{vis_NY} and Fig.\ref{vis_CNSatMap} also showcase EarthMapper’s ability to produce realistic satellite textures and precise map structures, highlighting its exceptional quality and fidelity in geospatial translation tasks. Besides, our EarthMapper demonstrates robust diversity in generated images, producing varied yet semantically consistent outputs from identical conditional inputs, as illustrated in Fig.~\ref{diverse}.

\begin{figure}[tbp]
	\begin{center}
		\centerline{\includegraphics[width=1\linewidth]{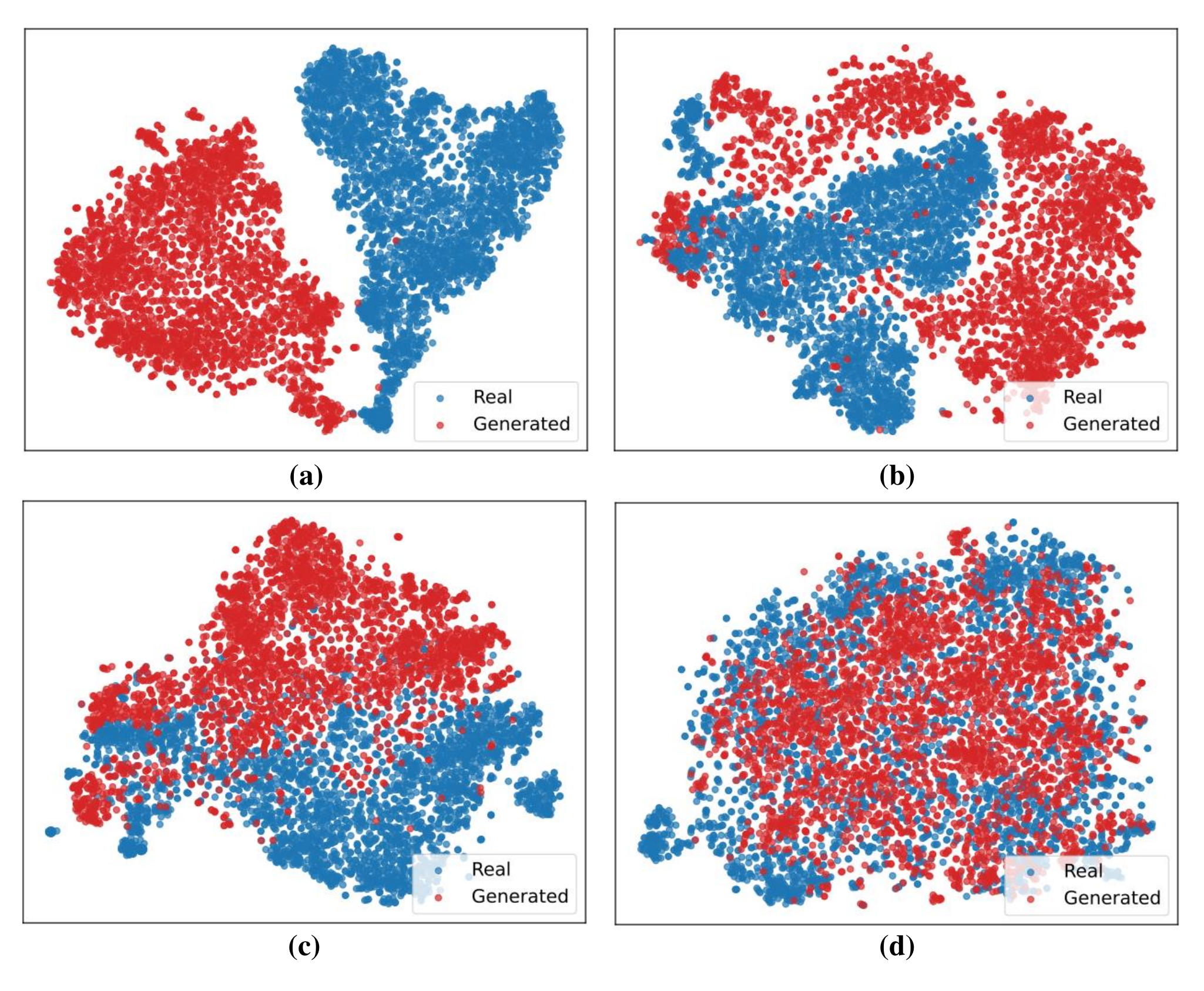}}
		
		\caption{T-SNE visualization comparing different methods on  the New York test set. The blue points represent the feature distribution of the ground truth satellite images, while the red points indicate the feature distribution of the generated satellite images. (a) BBDM. (b) Pix2Pix. (c) CAR. (d) EarthMapper.}\label{tsne}
	\end{center}
\end{figure}

Complementing these findings, the t-SNE visualization in Fig.~\ref{tsne} further demonstrates EarthMapper’s superior feature alignment. Generated samples (red) form a tight, cohesive cluster closely aligned with real satellite imagery (blue). In comparison, BBDM exhibits significant dispersion, Pix2Pix shows partial overlap with fragmented clustering, and CAR, while improved, lacks the same level of coherence. This visualization emphasizes EarthMapper’s capability to accurately capture the real data manifold, ensuring enhanced fidelity and structural consistency, setting it apart from other methods.

	\begin{table*}[t]
	\centering
	\caption{Ablation Study of Key Components in EarthMapper on the CNSatMap Test Set.}
	\label{tab:performance_comparison}
	
	\begin{tabular}{clcccc}
		\toprule \toprule
		\multirow{2}{*}{ID} & \multirow{2}{*}{Method} & \multicolumn{2}{c}{Map-to-Satellite} & \multicolumn{2}{c}{Satellite-to-Map} \\
		\cmidrule(lr){3-4} \cmidrule(lr){5-6}
		& & FID$\downarrow$ & KID$\downarrow$ & SSIM$\uparrow$ & PSNR$\uparrow$ \\
		\midrule
		1 & Baseline  & 64.28 & 5.64 & 0.5329 & 20.87 \\
		2 & + Geo-Conditioned Joint Scale Autoregression & 47.33 \textbf{\textcolor{green!50!black}{\scriptsize{($-$16.95)}}} & 4.51 \textbf{\textcolor{green!50!black}{\scriptsize{($-$1.13)}}} & 0.6381 \textbf{\textcolor{green!50!black}{\scriptsize{($+$0.1052)}}} & 23.46 \textbf{\textcolor{green!50!black}{\scriptsize{($+$2.59)}}} \\
		3 & + Semantic Infusion & 34.69 \textbf{\textcolor{green!50!black}{\scriptsize{($-$12.64)}}} & 3.92 \textbf{\textcolor{green!50!black}{\scriptsize{($-$0.59)}}} & 0.6629 \textbf{\textcolor{green!50!black}{\scriptsize{($+$0.0248)}}} & 25.39 \textbf{\textcolor{green!50!black}{\scriptsize{($+$1.93)}}}\\
		4 & + Key Point Force & 31.57\textbf{\textcolor{green!50!black}{\scriptsize{($-$3.12)}}} & 2.85 \textbf{\textcolor{green!50!black}{\scriptsize{($-$1.07)}}} & 0.7054 \textbf{\textcolor{green!50!black}{\scriptsize{($+$0.0425)}}} & 26.05 \textbf{\textcolor{green!50!black}{\scriptsize{($+$0.66)}}} \\
		5 & + Complexity Guidance & 29.89 \textbf{\textcolor{green!50!black}{\scriptsize{($-$1.68)}}} & 2.06 \textbf{\textcolor{green!50!black}{\scriptsize{($-$0.79)}}} & 0.7300 \textbf{\textcolor{green!50!black}{\scriptsize{($+$0.0246)}}} & 26.88 \textbf{\textcolor{green!50!black}{\scriptsize{($+$0.83)}}} \\
		\bottomrule \bottomrule
	\end{tabular}
\end{table*}

\subsection{Ablation Study}
\label{section:Ablation}

\subsubsection{Effectiveness of EarthMapper's Components}

The ablation study, detailed in Table~\ref{tab:performance_comparison}, systematically dissects the contributions of EarthMapper’s components by building upon a baseline defined as the VAR with joint modeling. Introducing geo-conditioned joint scale autoregression enhances the baseline by embedding geospatial priors into the autoregressive process, enabling the model to explicitly account for spatial relationships and scale variations inherent in remote sensing data. This conditioning aligns token predictions with geographic structures, substantially improving feature distribution coherence in map-to-satellite translation and structural preservation in satellite-to-map translation tasks, as evidenced by the marked enhancement in visual realism and reconstruction fidelity.

Further advancements are achieved through semantic infusion, key point force, and complexity guidance, each addressing distinct limitations of the baseline’s generic autoregressive approach. Semantic infusion integrates pretrained visual embeddings to enrich the model’s understanding of high-level semantics, fostering precise cross-modal alignment by bridging linguistic and visual representations. Key point force dynamically emphasizes critical spatial keypoints, adapting feature weighting to prioritize structurally significant regions, thus refining the model’s ability to reconstruct complex geospatial patterns. Finally, complexity guidance introduces a dynamic scaling mechanism that modulates conditional guidance based on image complexity, balancing generative diversity with reconstruction accuracy. This adaptive control mitigates overfitting to simplistic patterns, ensuring robust performance across diverse urban scenes. Together, these components transform the baseline’s generic framework into a specialized architecture adept at bidirectional geospatial translation, with each module addressing principled deficiencies to achieve superior alignment and fidelity.

\begin{table}[t]
	\centering
	\caption{Ablation Analysis of Guidance Scales for map-to-satellite Translation on the New York Dataset. The Best Performance Is Shown in Bold.}
	\label{cfg_map2sat}
	\renewcommand\arraystretch{1.0}
	\fontsize{8}{12}\selectfont
	\begin{tabular}{c|c|c|c|c}
		\hline \hline
		\textbf{Guidance Scale}  &  \textbf{FID $\downarrow$} & \textbf{KID $\downarrow$} & \textbf{Precision $\uparrow$} & \textbf{Recall $\uparrow$} \\ \hline
		{[}2, 2, 2{]} & 39.60 & 1.24 & 0.4279 & 0.4907 \\ 
		{[}4, 4, 4{]} & 39.92 & 1.20 & 0.4589 & \textbf{0.5099} \\ 
		{[}6, 6, 6{]} & 37.86 & 1.16 & 0.5620 & 0.4775 \\ 
		{[}8, 8, 8{]} & \textbf{36.54} & \textbf{0.99} & \textbf{0.6182} & 0.4890 \\ 
		{[}10, 10, 10{]} & 39.26 & 1.08 & 0.5748 & 0.4538 \\
		{[}12, 12, 12{]} & 42.39 & 1.35 & 0.5589 & 0.4215 \\
		
		\hline \hline
	\end{tabular}
\end{table}

\begin{table}[t]
	\centering
	\caption{Ablation Analysis of Guidance Scales for Satellite-to-Map Translation on the New York Dataset. The Best Performance Is Shown in Bold.}
	\label{cfg_sat2map}
	\renewcommand\arraystretch{1.0}
	\fontsize{8}{12}\selectfont
	\begin{tabular}{c|c|c|c|c}
		\hline \hline
		\textbf{Guidance Scale}  &  \textbf{SSIM} $\uparrow$ & \textbf{PSNR $\uparrow$} & \textbf{RMSE $\downarrow$} & \textbf{LPIPS $\downarrow$} \\ \hline
		{[}1, 1, 1{]} & 0.6363 & 24.48 & 0.0635 & 0.2969 \\
		{[}2, 2, 2{]} & \textbf{0.6534} & \textbf{25.04} & \textbf{0.0611} & \textbf{0.2819}  \\ 
		{[}4, 4, 4{]} & 0.6477 & 24.46 & 0.0638 & 0.2960  \\ 
		{[}6, 6, 6{]} & 0.6374 & 24.26 & 0.0639 & 0.2961  \\ 
		{[}8, 8, 8{]} & 0.6263 & 24.17 & 0.0640 & 0.2978  \\ 
		{[}10, 10, 10{]} & 0.6140 & 24.04 & 0.0644 & 0.2966  \\

		\hline \hline
	\end{tabular}
\end{table}

\subsubsection{The Impact of Different Guidance Scales}

To elucidate the role of the guidance scale in EarthMapper’s complexity guidance mechanism, we conducted an ablation study on the New York test set, evaluating its influence on BSMT task. The results, presented in Tables~\ref{cfg_map2sat} and Tables~\ref{cfg_sat2map}, systematically explore a range of guidance scales. For map-to-satellite translation, a guidance scale of [8, 8, 8] achieves optimal performance, yielding an FID of 36.54, KID of 0.99, Precision of 0.6182, and Recall of 0.4890. This configuration excels in balancing generative fidelity and diversity, as evidenced by the minimal distributional discrepancies (FID and KID) and high fidelity within the real image manifold (Precision). Lower scales, such as [2, 2, 2], result in higher FID (39.60) and KID (1.24), indicating insufficient conditional control, while higher scales, like [12, 12, 12], degrade performance (FID: 42.39, KID: 1.35), suggesting overfitting to conditional constraints that stifles generative diversity.

In contrast, for satellite-to-map translation, a guidance scale of [2, 2, 2] delivers the best performance, with an SSIM of 0.6534, PSNR of 25.04, RMSE of 0.0611, and LPIPS of 0.2819, reflecting superior structural preservation and perceptual similarity. Higher scales, such as [4, 4, 4] and beyond, progressively degrade performance (e.g., SSIM: 0.6140 at [10, 10, 10]), indicating excessive guidance that distorts fine-grained map structures. The divergence in optimal scales between tasks stems from their inherent objectives: map-to-satellite translation, an open-domain generation task, benefits from stronger guidance ([8, 8, 8]) to align complex, variable satellite textures with the real data manifold, requiring robust conditional steering to ensure realism. Conversely, for satellite-to-map translation, a structured reconstruction task, demands precise pixel-level accuracy and structural fidelity, where lighter guidance ([2, 2, 2]) prevents over-constraining the model, preserving intricate cartographic details.

\begin{figure*}[tbp]
	\begin{center}
		\centerline{\includegraphics[width=1\linewidth]{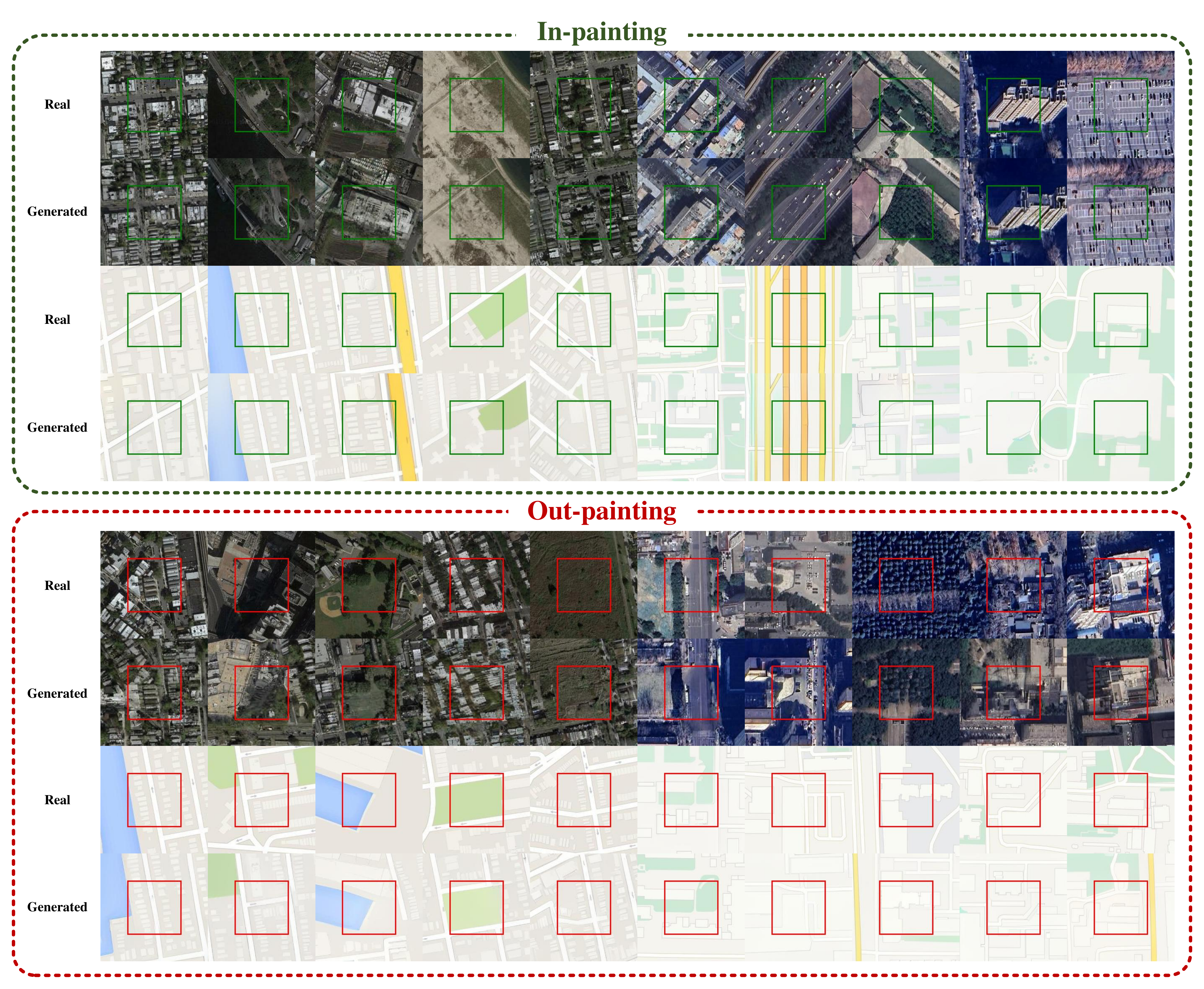}}
		
		\caption{Zero-shot performance of EarthMapper in in-painting and out-painting tasks.}\label{zero-shot}
		\vspace{-10pt} 
	\end{center}
\end{figure*}

\begin{table}[t]
	\centering
	\caption{Performance Evaluation of Cross-Dataset Generalization for Bidirectional Satellite-Map Translation}
	\renewcommand\arraystretch{2.0}
	\label{cross-dataset}
	\begin{tabular}{c|c|c|c|c}
		\hline \hline
		\multirow{2}{*}{Cross-dataset Generalization} & \multicolumn{2}{c|}{Map-to-Satellite} & \multicolumn{2}{c}{Satellite-to-Map} \\
		\cline{2-5}
		& FID $\downarrow$ & KID $\downarrow$ & SSIM $\uparrow$ & PSNR $\uparrow$ \\
		\hline
		CNSatMap$\rightarrow$New York & \textbf{164.91} & \textbf{17.68} & \textbf{0.5226} & \textbf{22.50} \\
		\hline
		New York$\rightarrow$CNSatMap & 236.83 & 20.84 & 0.2711 & 8.18 \\
		\hline \hline
	\end{tabular}
\end{table}

\subsection{Zero-shot Generalization}
\label{section:Zero-shot}

\subsubsection{Cross-dataset Generalization}

To evaluate the robustness and generalizability of the proposed EarthMapper framework across diverse geographical contexts, we conducted a cross-dataset generalization experiment, with results presented in Table~\ref{cross-dataset}. The experiment assesses BSMT performance by training EarthMapper on one dataset and testing it on another, specifically from CNSatMap to New York and vice versa. For the map-to-satellite translation, EarthMapper trained on CNSatMap and tested on New York achieves an FID of 164.91 and KID of 17.68, significantly outperforming the reverse configuration (New York$\rightarrow$CNSatMap), which yields an FID of 236.83 and KID of 20.84. Similarly, in satellite-to-map translation, the CNSatMap$\rightarrow$New York setup delivers superior results with an SSIM of 0.5226 and PSNR of 22.50, compared to 0.2711 and 8.18 for New York$\rightarrow$CNSatMap. These results underscore the CNSatMap dataset's exceptional capacity to foster robust generalization. The dataset’s extensive scale, encompassing 302,132 high-quality satellite-map pairs across 38 diverse Chinese cities, captures a broad spectrum of urban and topographical variations, enabling models to learn rich, transferable representations. In contrast, the New York dataset, while valuable, is limited in geographical diversity, leading to overfitting and poor generalization when applied to the more varied CNSatMap test set.

\subsubsection{In-painting and Out-painting}

To rigorously evaluate EarthMapper’s generalization to unseen geospatial distributions, we assess its zero-shot performance on in-painting (reconstructing occluded regions) and out-painting (extending beyond image boundaries)—tasks requiring robust contextual reasoning and spatial coherence. As illustrated in Fig.~\ref{zero-shot}, EarthMapper reconstructs occluded regions, such as obscured buildings and roads, with high fidelity, seamlessly integrating synthesized textures with existing structures. Notably, EarthMapper avoids common artifacts, including blurring and misalignment, even under severe occlusion, demonstrating its ability to generalize beyond training distributions. For out-painting, EarthMapper generates realistic urban layouts and natural topographies without geometric distortions. Crucially, this performance is achieved without fine-tuning, highlighting EarthMapper’s inherent adaptability to diverse geospatial contexts.

\subsubsection{Coordinate-conditional Image Generation}

\begin{figure}[tbp]
	\begin{center}
		\centerline{\includegraphics[width=1\linewidth]{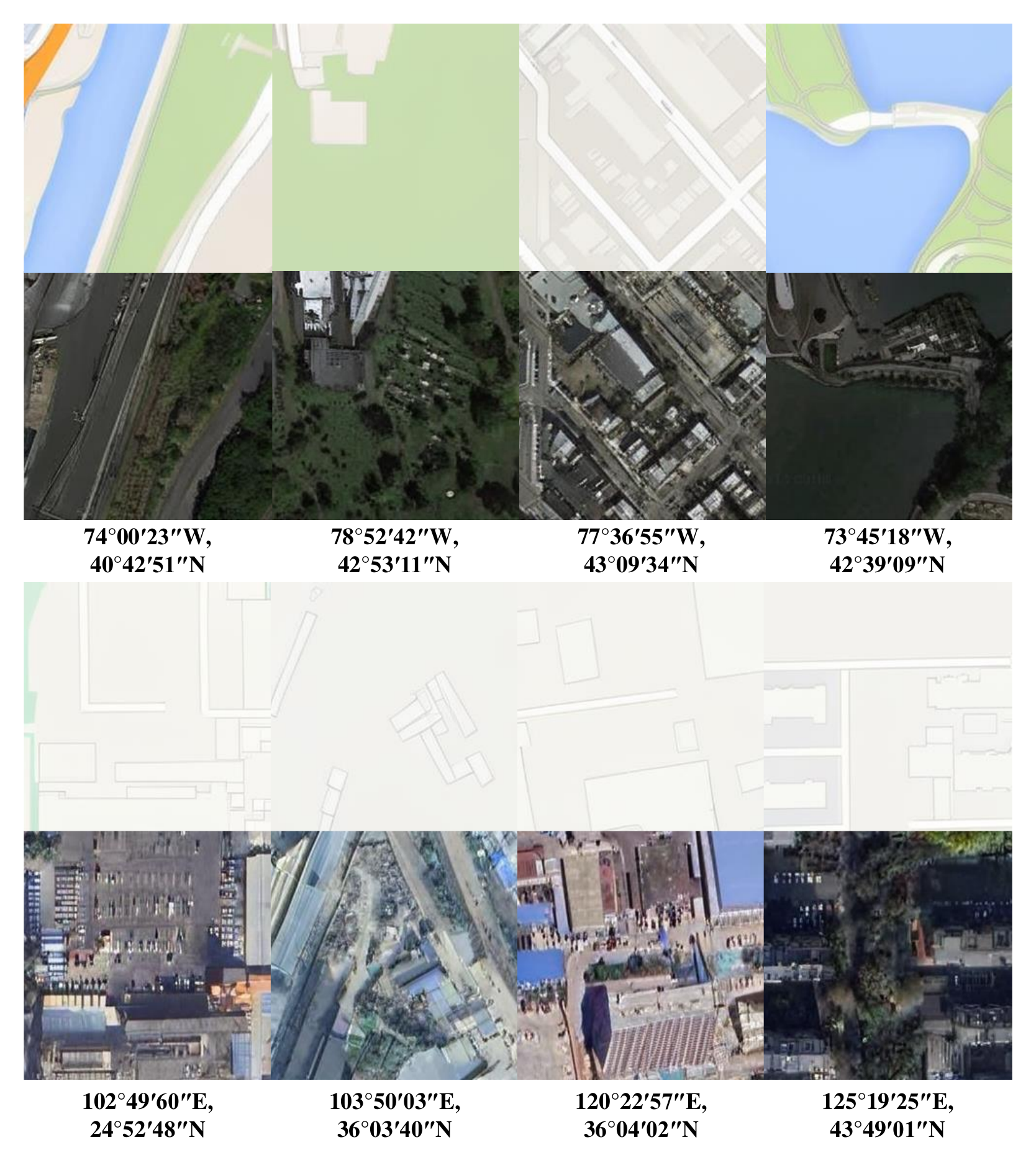}}
		
		\caption{Zero-shot performance of EarthMapper in coordinate-conditional image generation task.}\label{coord_gen}
		\vspace{-20pt} 
	\end{center}
\end{figure}

To further evaluate the generalization capabilities of EarthMapper, we conduct a zero-shot coordinate-conditional image generation experiment. This task assesses the model's ability to infer complex geospatial patterns without additional contextual inputs. As illustrated in Fig.~\ref{coord_gen}, EarthMapper generates high-fidelity satellite-map image pairs that accurately capture the spatial and semantic characteristics of the provided coordinates, including urban layouts and natural features. The generated images demonstrate strong consistency between satellite and map representations, exhibiting coherent textures and structures. These results underscore EarthMapper's robust generalization across diverse geospatial contexts, achieved without fine-tuning or auxiliary data.

\section{Conclusion}
\label{section:Conclusion}

In this paper, we present EarthMapper, an innovative autoregressive framework that seamlessly integrates geographic coordinate embeddings with multi-scale feature alignment to achieve high-fidelity bidirectional satellite-map translation. This approach employs geo-conditioned joint scale autoregression (GJSA), enhanced by semantic infusion (SI) for feature consistency and key point adaptive guidance (KPAG) for balanced diversity and precision. We also construct CNSatMap, a dataset of 302,132 aligned satellite-map pairs across 38 Chinese cities, providing a robust benchmark for cross-modal research. Evaluations on CNSatMap and New York datasets show superior performance over state-of-the-art methods, with enhanced visual realism, semantic consistency, and structural fidelity. The framework’s versatility in zero-shot tasks, including in-painting, out-painting, and coordinate-conditional generation, highlights its strong generalization across diverse geospatial contexts.

The significance of this work lies in its ability to bridge the visual-semantic gap between raw satellite data and human-interpretable maps, offering a scalable solution that balances precision and diversity. While EarthMapper sets a new standard in bidirectional translation, several avenues remain for future exploration. Integrating additional modalities, such as LiDAR or hyperspectral data, could further enrich translation quality by capturing complementary geospatial attributes. Extending EarthMapper to tasks like change detection or land cover classification may broaden its applicability in remote sensing. Moreover, optimizing its computational efficiency could enable real-time deployment, enhancing its practical utility. Collectively, these directions promise to amplify EarthMapper’s impact, paving the way for next-generation geospatial intelligence.

%

\ifCLASSOPTIONcaptionsoff
\newpage
\fi
\bibliography{refs}

\begin{thebibliography}{10}
\providecommand{\url}[1]{#1}
\csname url@samestyle\endcsname
\providecommand{\newblock}{\relax}
\providecommand{\bibinfo}[2]{#2}
\providecommand{\BIBentrySTDinterwordspacing}{\spaceskip=0pt\relax}
\providecommand{\BIBentryALTinterwordstretchfactor}{4}
\providecommand{\BIBentryALTinterwordspacing}{\spaceskip=\fontdimen2\font plus
\BIBentryALTinterwordstretchfactor\fontdimen3\font minus
  \fontdimen4\font\relax}
\providecommand{\BIBforeignlanguage}[2]{{%
\expandafter\ifx\csname l@#1\endcsname\relax
\typeout{** WARNING: IEEEtran.bst: No hyphenation pattern has been}%
\typeout{** loaded for the language `#1'. Using the pattern for}%
\typeout{** the default language instead.}%
\else
\language=\csname l@#1\endcsname
\fi
#2}}
\providecommand{\BIBdecl}{\relax}
\BIBdecl

\bibitem{ingale2021image}
V.~Ingale, R.~Singh, and P.~Patwal, ``Image to image translation: Generating
  maps from satellite images,'' \emph{arXiv preprint arXiv:2105.09253}, 2021.

\bibitem{espinosa2023generate}
M.~Espinosa and E.~J. Crowley, ``Generate your own scotland: Satellite image
  generation conditioned on maps,'' \emph{arXiv preprint arXiv:2308.16648},
  2023.

\bibitem{goodfellow2020generative}
I.~Goodfellow, J.~Pouget-Abadie, M.~Mirza, B.~Xu, D.~Warde-Farley, S.~Ozair,
  A.~Courville, and Y.~Bengio, ``Generative adversarial networks,''
  \emph{Communications of the ACM}, vol.~63, no.~11, pp. 139--144, 2020.

\bibitem{mirza2014conditional}
M.~Mirza and S.~Osindero, ``Conditional generative adversarial nets,''
  \emph{arXiv preprint arXiv:1411.1784}, 2014.

\bibitem{isola2017image}
P.~Isola, J.-Y. Zhu, T.~Zhou, and A.~A. Efros, ``Image-to-image translation
  with conditional adversarial networks,'' in \emph{Proceedings of the IEEE
  conference on computer vision and pattern recognition}, 2017, pp. 1125--1134.

\bibitem{zhu2017unpaired}
J.-Y. Zhu, T.~Park, P.~Isola, and A.~A. Efros, ``Unpaired image-to-image
  translation using cycle-consistent adversarial networks,'' in
  \emph{Proceedings of the IEEE international conference on computer vision},
  2017, pp. 2223--2232.

\bibitem{ho2020denoising}
J.~Ho, A.~Jain, and P.~Abbeel, ``Denoising diffusion probabilistic models,''
  \emph{Advances in neural information processing systems}, vol.~33, pp.
  6840--6851, 2020.

\bibitem{dhariwal2021diffusion}
P.~Dhariwal and A.~Nichol, ``Diffusion models beat gans on image synthesis,''
  \emph{Advances in neural information processing systems}, vol.~34, pp.
  8780--8794, 2021.

\bibitem{chen2020generative}
M.~Chen, A.~Radford, R.~Child, J.~Wu, H.~Jun, D.~Luan, and I.~Sutskever,
  ``Generative pretraining from pixels,'' in \emph{International conference on
  machine learning}.\hskip 1em plus 0.5em minus 0.4em\relax PMLR, 2020, pp.
  1691--1703.

\bibitem{van2016conditional}
A.~Van~den Oord, N.~Kalchbrenner, L.~Espeholt, O.~Vinyals, A.~Graves
  \emph{et~al.}, ``Conditional image generation with pixelcnn decoders,''
  \emph{Advances in neural information processing systems}, vol.~29, 2016.

\bibitem{tian2024visual}
K.~Tian, Y.~Jiang, Z.~Yuan, B.~Peng, and L.~Wang, ``Visual autoregressive
  modeling: Scalable image generation via next-scale prediction,''
  \emph{Advances in neural information processing systems}, vol.~37, pp.
  84\,839--84\,865, 2024.

\bibitem{li2024controlvar}
X.~Li, K.~Qiu, H.~Chen, J.~Kuen, Z.~Lin, R.~Singh, and B.~Raj, ``Controlvar:
  Exploring controllable visual autoregressive modeling,'' \emph{arXiv preprint
  arXiv:2406.09750}, 2024.

\bibitem{yao2024car}
Z.~Yao, J.~Li, Y.~Zhou, Y.~Liu, X.~Jiang, C.~Wang, F.~Zheng, Y.~Zou, and L.~Li,
  ``Car: Controllable autoregressive modeling for visual generation,''
  \emph{arXiv preprint arXiv:2410.04671}, 2024.

\bibitem{zhang2023adding}
L.~Zhang, A.~Rao, and M.~Agrawala, ``Adding conditional control to
  text-to-image diffusion models,'' in \emph{Proceedings of the IEEE/CVF
  international conference on computer vision}, 2023, pp. 3836--3847.

\bibitem{rombach2022high}
R.~Rombach, A.~Blattmann, D.~Lorenz, P.~Esser, and B.~Ommer, ``High-resolution
  image synthesis with latent diffusion models,'' in \emph{Proceedings of the
  IEEE/CVF conference on computer vision and pattern recognition}, 2022, pp.
  10\,684--10\,695.

\bibitem{li2023gligen}
Y.~Li, H.~Liu, Q.~Wu, F.~Mu, J.~Yang, J.~Gao, C.~Li, and Y.~J. Lee, ``Gligen:
  Open-set grounded text-to-image generation,'' in \emph{Proceedings of the
  IEEE/CVF conference on computer vision and pattern recognition}, 2023, pp.
  22\,511--22\,521.

\bibitem{karras2019style}
T.~Karras, S.~Laine, and T.~Aila, ``A style-based generator architecture for
  generative adversarial networks,'' in \emph{Proceedings of the IEEE/CVF
  conference on computer vision and pattern recognition}, 2019, pp. 4401--4410.

\bibitem{shen2020interfacegan}
Y.~Shen, C.~Yang, X.~Tang, and B.~Zhou, ``Interfacegan: Interpreting the
  disentangled face representation learned by gans,'' \emph{IEEE transactions
  on pattern analysis and machine intelligence}, vol.~44, no.~4, pp.
  2004--2018, 2020.

\bibitem{takagi2023high}
Y.~Takagi and S.~Nishimoto, ``High-resolution image reconstruction with latent
  diffusion models from human brain activity,'' in \emph{Proceedings of the
  IEEE/CVF Conference on Computer Vision and Pattern Recognition}, 2023, pp.
  14\,453--14\,463.

\bibitem{gal2022image}
R.~Gal, Y.~Alaluf, Y.~Atzmon, O.~Patashnik, A.~H. Bermano, G.~Chechik, and
  D.~Cohen-Or, ``An image is worth one word: Personalizing text-to-image
  generation using textual inversion,'' \emph{arXiv preprint arXiv:2208.01618},
  2022.

\bibitem{ruiz2023dreambooth}
N.~Ruiz, Y.~Li, V.~Jampani, Y.~Pritch, M.~Rubinstein, and K.~Aberman,
  ``Dreambooth: Fine tuning text-to-image diffusion models for subject-driven
  generation,'' in \emph{Proceedings of the IEEE/CVF conference on computer
  vision and pattern recognition}, 2023, pp. 22\,500--22\,510.

\bibitem{nichol2021glide}
A.~Nichol, P.~Dhariwal, A.~Ramesh, P.~Shyam, P.~Mishkin, B.~McGrew,
  I.~Sutskever, and M.~Chen, ``Glide: Towards photorealistic image generation
  and editing with text-guided diffusion models,'' \emph{arXiv preprint
  arXiv:2112.10741}, 2021.

\bibitem{ramesh2022hierarchical}
A.~Ramesh, P.~Dhariwal, A.~Nichol, C.~Chu, and M.~Chen, ``Hierarchical
  text-conditional image generation with clip latents,'' \emph{arXiv preprint
  arXiv:2204.06125}, vol.~1, no.~2, p.~3, 2022.

\bibitem{radford2021learning}
A.~Radford, J.~W. Kim, C.~Hallacy, A.~Ramesh, G.~Goh, S.~Agarwal, G.~Sastry,
  A.~Askell, P.~Mishkin, J.~Clark \emph{et~al.}, ``Learning transferable visual
  models from natural language supervision,'' in \emph{International conference
  on machine learning}.\hskip 1em plus 0.5em minus 0.4em\relax PmLR, 2021, pp.
  8748--8763.

\bibitem{ho2022classifier}
J.~Ho and T.~Salimans, ``Classifier-free diffusion guidance,'' \emph{arXiv
  preprint arXiv:2207.12598}, 2022.

\bibitem{tang2023any}
Z.~Tang, Z.~Yang, C.~Zhu, M.~Zeng, and M.~Bansal, ``Any-to-any generation via
  composable diffusion,'' \emph{Advances in Neural Information Processing
  Systems}, vol.~36, pp. 16\,083--16\,099, 2023.

\bibitem{radford2019language}
A.~Radford, J.~Wu, R.~Child, D.~Luan, D.~Amodei, I.~Sutskever \emph{et~al.},
  ``Language models are unsupervised multitask learners,'' \emph{OpenAI blog},
  vol.~1, no.~8, p.~9, 2019.

\bibitem{brown2020language}
T.~Brown, B.~Mann, N.~Ryder, M.~Subbiah, J.~D. Kaplan, P.~Dhariwal,
  A.~Neelakantan, P.~Shyam, G.~Sastry, A.~Askell \emph{et~al.}, ``Language
  models are few-shot learners,'' \emph{Advances in neural information
  processing systems}, vol.~33, pp. 1877--1901, 2020.

\bibitem{parmar2018image}
N.~Parmar, A.~Vaswani, J.~Uszkoreit, L.~Kaiser, N.~Shazeer, A.~Ku, and D.~Tran,
  ``Image transformer,'' in \emph{International conference on machine
  learning}.\hskip 1em plus 0.5em minus 0.4em\relax PMLR, 2018, pp. 4055--4064.

\bibitem{li2016video}
K.~Li, Y.~Zhu, J.~Yang, and J.~Jiang, ``Video super-resolution using an
  adaptive superpixel-guided auto-regressive model,'' \emph{Pattern
  Recognition}, vol.~51, pp. 59--71, 2016.

\bibitem{guo2022lar}
B.~Guo, X.~Zhang, H.~Wu, Y.~Wang, Y.~Zhang, and Y.-F. Wang, ``Lar-sr: A local
  autoregressive model for image super-resolution,'' in \emph{Proceedings of
  the IEEE/CVF conference on computer vision and pattern recognition}, 2022,
  pp. 1909--1918.

\bibitem{yao2022outpainting}
K.~Yao, P.~Gao, X.~Yang, J.~Sun, R.~Zhang, and K.~Huang, ``Outpainting by
  queries,'' in \emph{European conference on computer vision}.\hskip 1em plus
  0.5em minus 0.4em\relax Springer, 2022, pp. 153--169.

\bibitem{crowson2022vqgan}
K.~Crowson, S.~Biderman, D.~Kornis, D.~Stander, E.~Hallahan, L.~Castricato, and
  E.~Raff, ``Vqgan-clip: Open domain image generation and editing with natural
  language guidance,'' in \emph{European conference on computer vision}.\hskip
  1em plus 0.5em minus 0.4em\relax Springer, 2022, pp. 88--105.

\bibitem{li2024controlar}
Z.~Li, T.~Cheng, S.~Chen, P.~Sun, H.~Shen, L.~Ran, X.~Chen, W.~Liu, and
  X.~Wang, ``Controlar: Controllable image generation with autoregressive
  models,'' \emph{arXiv preprint arXiv:2410.02705}, 2024.

\bibitem{van2016pixel}
A.~Van Den~Oord, N.~Kalchbrenner, and K.~Kavukcuoglu, ``Pixel recurrent neural
  networks,'' in \emph{International conference on machine learning}.\hskip 1em
  plus 0.5em minus 0.4em\relax PMLR, 2016, pp. 1747--1756.

\bibitem{reed2017parallel}
S.~Reed, A.~Oord, N.~Kalchbrenner, S.~G. Colmenarejo, Z.~Wang, Y.~Chen,
  D.~Belov, and N.~Freitas, ``Parallel multiscale autoregressive density
  estimation,'' in \emph{International conference on machine learning}.\hskip
  1em plus 0.5em minus 0.4em\relax PMLR, 2017, pp. 2912--2921.

\bibitem{van2017neural}
A.~Van Den~Oord, O.~Vinyals \emph{et~al.}, ``Neural discrete representation
  learning,'' \emph{Advances in neural information processing systems},
  vol.~30, 2017.

\bibitem{razavi2019generating}
A.~Razavi, A.~Van~den Oord, and O.~Vinyals, ``Generating diverse high-fidelity
  images with vq-vae-2,'' \emph{Advances in neural information processing
  systems}, vol.~32, 2019.

\bibitem{esser2021taming}
P.~Esser, R.~Rombach, and B.~Ommer, ``Taming transformers for high-resolution
  image synthesis,'' in \emph{Proceedings of the IEEE/CVF conference on
  computer vision and pattern recognition}, 2021, pp. 12\,873--12\,883.

\bibitem{vaswani2017attention}
A.~Vaswani, N.~Shazeer, N.~Parmar, J.~Uszkoreit, L.~Jones, A.~N. Gomez,
  {\L}.~Kaiser, and I.~Polosukhin, ``Attention is all you need,''
  \emph{Advances in neural information processing systems}, vol.~30, 2017.

\bibitem{lee2022autoregressive}
D.~Lee, C.~Kim, S.~Kim, M.~Cho, and W.-S. Han, ``Autoregressive image
  generation using residual quantization,'' in \emph{Proceedings of the
  IEEE/CVF Conference on Computer Vision and Pattern Recognition}, 2022, pp.
  11\,523--11\,532.

\bibitem{pang2024hsigene}
L.~Pang, X.~Cao, D.~Tang, S.~Xu, X.~Bai, F.~Zhou, and D.~Meng, ``Hsigene: A
  foundation model for hyperspectral image generation,'' \emph{arXiv preprint
  arXiv:2409.12470}, 2024.

\bibitem{sebaq2024rsdiff}
A.~Sebaq and M.~ElHelw, ``Rsdiff: Remote sensing image generation from text
  using diffusion model,'' \emph{Neural Computing and Applications}, vol.~36,
  no.~36, pp. 23\,103--23\,111, 2024.

\bibitem{tang2024crs}
D.~Tang, X.~Cao, X.~Hou, Z.~Jiang, and D.~Meng, ``Crs-diff: Controllable
  generative remote sensing foundation model,'' \emph{arXiv e-prints}, pp.
  arXiv--2403, 2024.

\bibitem{yu2024metaearth}
Z.~Yu, C.~Liu, L.~Liu, Z.~Shi, and Z.~Zou, ``Metaearth: A generative foundation
  model for global-scale remote sensing image generation,'' \emph{IEEE
  Transactions on Pattern Analysis and Machine Intelligence}, 2024.

\bibitem{khanna2023diffusionsat}
S.~Khanna, P.~Liu, L.~Zhou, C.~Meng, R.~Rombach, M.~Burke, D.~Lobell, and
  S.~Ermon, ``Diffusionsat: A generative foundation model for satellite
  imagery,'' \emph{arXiv preprint arXiv:2312.03606}, 2023.

\bibitem{heusel2017gans}
M.~Heusel, H.~Ramsauer, T.~Unterthiner, B.~Nessler, and S.~Hochreiter, ``Gans
  trained by a two time-scale update rule converge to a local nash
  equilibrium,'' \emph{Advances in neural information processing systems},
  vol.~30, 2017.

\bibitem{binkowski2018demystifying}
M.~Bi{\'n}kowski, D.~J. Sutherland, M.~Arbel, and A.~Gretton, ``Demystifying
  mmd gans,'' \emph{arXiv preprint arXiv:1801.01401}, 2018.

\bibitem{kynkaanniemi2019improved}
T.~Kynk{\"a}{\"a}nniemi, T.~Karras, S.~Laine, J.~Lehtinen, and T.~Aila,
  ``Improved precision and recall metric for assessing generative models,''
  \emph{Advances in neural information processing systems}, vol.~32, 2019.

\bibitem{zhang2018unreasonable}
R.~Zhang, P.~Isola, A.~A. Efros, E.~Shechtman, and O.~Wang, ``The unreasonable
  effectiveness of deep features as a perceptual metric,'' in \emph{Proceedings
  of the IEEE conference on computer vision and pattern recognition}, 2018, pp.
  586--595.

\bibitem{oquab2023dinov2}
M.~Oquab, T.~Darcet, T.~Moutakanni, H.~Vo, M.~Szafraniec, V.~Khalidov,
  P.~Fernandez, D.~Haziza, F.~Massa, A.~El-Nouby \emph{et~al.}, ``Dinov2:
  Learning robust visual features without supervision,'' \emph{arXiv preprint
  arXiv:2304.07193}, 2023.

\bibitem{loshchilov2017decoupled}
I.~Loshchilov and F.~Hutter, ``Decoupled weight decay regularization,''
  \emph{arXiv preprint arXiv:1711.05101}, 2017.

\bibitem{wu2024stegogan}
S.~Wu, Y.~Chen, S.~Mermet, L.~Hurni, K.~Schindler, N.~Gonthier, and
  L.~Landrieu, ``Stegogan: Leveraging steganography for non-bijective
  image-to-image translation,'' in \emph{Proceedings of the IEEE/CVF Conference
  on Computer Vision and Pattern Recognition}, 2024, pp. 7922--7931.

\bibitem{li2023bbdm}
B.~Li, K.~Xue, B.~Liu, and Y.-K. Lai, ``Bbdm: Image-to-image translation with
  brownian bridge diffusion models,'' in \emph{Proceedings of the IEEE/CVF
  conference on computer vision and pattern Recognition}, 2023, pp. 1952--1961.

\bibitem{qin2023unicontrol}
C.~Qin, S.~Zhang, N.~Yu, Y.~Feng, X.~Yang, Y.~Zhou, H.~Wang, J.~C. Niebles,
  C.~Xiong, S.~Savarese \emph{et~al.}, ``Unicontrol: A unified diffusion model
  for controllable visual generation in the wild,'' \emph{arXiv preprint
  arXiv:2305.11147}, 2023.

\end{thebibliography}

\end{document}